\documentclass[3p,authoryear,a4paper]{elsarticle}

\usepackage{amsmath} % extensions for writing mathematical formulae
\usepackage{amssymb} % for math symbols
\usepackage[hidelinks]{hyperref} % hyperlink for in-text references
\usepackage{setspace}
\usepackage{graphicx} % to include images
\usepackage{amsfonts}
\usepackage{algorithm}
\usepackage{algorithmic}
\usepackage{array}
\usepackage{booktabs}
\usepackage{tikz}
\usepackage{tikz-cd}
\usepackage{pgfplots}
\usepackage{pgf}
\usepackage{lmodern}
\makeatletter
  \everymath=\expandafter{\the\everymath\displaystyle}%
\makeatother

\usepackage{newtxtext,newtxmath} 
\usepackage[capitalise,nameinlink]{cleveref}
\usepackage{float}

\begin{document}

% Title
\title{An Interpretable Deep Learning Model for General Insurance Pricing}
\author[unsw]{Patrick J. Laub}

\author[unsw]{Tu Pho\corref{cor1}}
\ead{tupho289@gmail.com}

\author[unsw]{Bernard Wong}
\cortext[cor1]{Corresponding author}

\address[unsw]{School of Risk and Actuarial Studies, UNSW Sydney NSW 2052, Australia}

% Abstract
\begin{abstract}

The rapid advancement of machine learning has provided an opportunity to transform the modeling techniques in actuarial analytics. Novel machine learning methods, especially deep learning, have demonstrated versatile modeling capability and superior predictive performance compared to traditional actuarial approaches such as Generalized Linear Models. However, the widespread adoption of deep learning techniques in the insurance industry is often hindered by the lack of model interpretability, as the intricacies of their inner workings remain obscured behind the complex model architecture. This lack of interpretability is further complicated by the absence of a generally accepted definition of what an interpretable model is. There are also various practical requirements, such as smoothness and monotonicity, that a pricing model in general insurance should possess in addition to being interpretable.

This paper introduces the Actuarial Neural Additive Model, an inherently interpretable deep learning model for general insurance pricing that offers fully transparent and interpretable results while retaining the strong predictive power of neural networks. This model assigns a dedicated neural network (or subnetwork) to each individual covariate and pairwise interaction term to independently learn its impact on the modeled output while implementing various architectural constraints to allow for essential interpretability (e.g. sparsity) and practical requirements (e.g. smoothness, monotonicity) in insurance applications. The development of our model is grounded in a solid foundation, where we establish a concrete definition of interpretability within the insurance context, complemented by a rigorous mathematical framework. Comparisons in terms of prediction accuracy are made with traditional actuarial and state-of-the-art machine learning methods using both synthetic and real insurance datasets. The results show that the proposed model outperforms other methods in most cases while offering complete transparency in its internal logic, underscoring the strong interpretability and predictive capability.

\end{abstract}

% Keywords (elsarticle format)
\begin{keyword}
General insurance \sep Deep learning \sep Interpretability \sep Neural additive models \sep Actuarial modeling
\end{keyword}

\maketitle

\section{Introduction}

\subsection{Background}
The most popular statistical model used in modeling general insurance claims is the Generalized Linear Model (GLM), introduced by \citet{nelder1972}. GLMs allow actuaries to incorporate a wide range of statistical distributions that are commonly adopted in actuarial analytics, and the underlying linearity assumption provides an explainable framework for claims modeling \citep{wuthrich2023a}. This model has been shown to work well in practice; however, deep learning---the subset of machine learning focusing on artificial neural network models---has been gaining substantial ground in recent years.

Applications of deep learning and other novel machine learning techniques in claim modeling have shown an improvement in prediction accuracy compared to classical methods such as the GLM \citep{noll2020, wuthrich2023b}. Nevertheless, the integration of such advanced techniques as the primary pricing method among actuaries has been slow since they are often perceived as ``black boxes'', where the intricacies of the inner workings remain obscured, making it challenging to decipher the rationale behind the models' predictions \citep{harris2024}. Although a key factor determining the success of an insurance company lies in its ability to accurately estimate future claim costs, there are various parties (e.g.\ shareholders, regulators, customers, reinsurers) whose interests are affected by the insurance premiums \citep{iaa2010}. Hence, it is crucial that the pricing process be transparent and ethical in addition to having a strong predictive power.

Various proposals in the actuarial literature have aimed to develop interpretable models incorporating deep learning components \citep{almudafer2022, avanzi2024a, mcdonnell2023, richman2022, richman2023a}. The most common approach enhances classical actuarial models such as the GLM by combining with neural networks, leveraging the interpretability of the classical methods in some sense. However, there remains a gap in this literature for a holistic approach developed based on a formal framework with a concrete definition of interpretability. % of interpreHowever, as no formal definition of interpretability exist the current gap in contemporary literature lies in the development of an inherently interpretable deep learning model that is built on a concrete definition of interpretability with well-defined constraints and practical requirements. Methods like LocalGLMnet \citep{richman2023a}, CAXNN \citep{richman2022}, or TabNet \citep{mcdonnell2023} are not developed based on a solid interpretability framework, which makes the resulting models contain certain aspects that are not fully interpretable or desirable in insurance pricing. 

In general, having a universal solution to the interpretability problem comes with various difficulties. Firstly, interpretability often depends on the context in which a model is applied \citep{rudin2022}. The criteria for an image recognition model, which processes pixels, may differ from those of an insurance pricing model that mainly uses tabular data. More generally speaking, there is currently a lack of discussion and consensus on the definition of interpretability and the requirements that an insurance pricing model should comply with to be considered interpretable \citep{delcaillau2022, murdoch2019}. Secondly, a model should be developed on a solid mathematical ground, and there is a challenge in how to define and allow for interpretability mathematically in such framework.

In the machine learning literature, a promising set of candidates for interpretable deep learning models is that of the  Neural Additive Model (NAM), first proposed by \citet{agarwal2021}, and its variations \citep{kraus2024, xu2023, yang2021}. These models assign each individual covariate and pairwise interaction effect with a dedicated neural network (or subnetwork) to independently learn its impact on the final outputs. This additive decomposition facilitates visualization and provides exact descriptions of the internal logic and results---allowing the model to be "inherently interpretable" \citep{agarwal2021}. Additionally, since these models are based on neural networks, they can be easily extended to allow for multi-task and multi-label learning, accommodate advanced modeling methods (e.g.\ entity embedding with high-cardinality categorical features), and handle various data types. However, the NAM structure is still fully flexible otherwise, and hence modifications to allow for additional features important to practical applications will be required before it can be suitable for actuarial use.

\subsection{Contributions}
In light of the aforementioned gaps and challenges, this paper presents two main contributions.
\begin{enumerate}
    \item Establish an interpretability framework in insurance pricing with a concrete definition and mathematical formulation of interpretable pricing models. We will describe how an interpretable model can be mathematically formulated in a supervised learning context and provide the interpretability and practical requirements that are important in the insurance pricing context.
    \item Develop the Actuarial Neural Additive Model (ANAM), an inherently interpretable model with neural network components that contains the essential features related to interpretability and other practical requirements in general insurance pricing. This is done by combining the Neural Additive Model with various enhancements (sparsity, smoothness, and monotonicity) in the architecture and training process to facilitate adoption in actuarial and insurance applications.
\end{enumerate}

\subsection{Outline of Paper}
The remainder of this paper is structured as follows. \Cref{sec:if} introduces the notations used in the paper and establishes the interpretability framework by discussing and specifying the definition of interpretability, the mathematical formulation of an inherently interpretable model in supervised learning, and the insurance pricing-specific interpretability requirements. \Cref{sec:mf} presents the architecture of ANAM. \Cref{sec:da} applies the proposed model to both synthetic and real datasets and compares its predictive performance with classical actuarial and black-box machine learning models. \Cref{sec:con} concludes by summarizing the contributions and results of the paper as well as outlining the current limitations and avenues for further research.

\section{Interpretability} \label{sec:if}

To create interpretable models, a fundamental question one needs to answer is what constitutes model interpretability.
\Cref{subsec:existing-defs-interpretability} outlines the disparate attempts to define this concept in the literature and describes traditional approaches to obtaining interpretable models.
\Cref{subsec:interpretability} gives our attempt to synthesize these past ideas into a concrete definition for the actuarial context.

\subsection{Existing Interpretability Definitions} \label{subsec:existing-defs-interpretability}
\subsubsection{Alternative Definitions}
\label{subsubsec:alternative-definitions}
Despite extensive research, interpretability in machine learning remains without a universally agreed definition \citep{delcaillau2022, murdoch2019}. A commonly cited view by \citet{doshi2017} defines interpretability as ``the ability to explain or to present in understandable terms to a human,'' emphasizing its human-centric nature. Similarly, \citet{gilpin2018} highlights interpretability as generating explanations simple enough to be comprehensible by a person using meaningful vocabulary.
However, these broad definitions face criticism due to their inherent subjectivity and broad-brush nature, potentially leading to disagreements about a model's interpretability \citep{li2022}. For instance, a neural network's general operation is explainable and well-understood, but it is widely considered a ``black box'' due to the complexity of its learned representations. This suggests that a definition should be thorough enough to allow for a concrete determination of interpretability.

Addressing these shortcomings, \citet{thampi2022} reframes interpretability through a lens of ``cause and effect,'' emphasizing the ability to predict model outcomes consistently, understand prediction reasoning, observe prediction sensitivity to inputs or parameters, and identify model errors. This perspective underscores the necessity for transparent internal logic and causality, which are also emphasized in \citet{gilpin2018}, \citet{miller2019}, and \citet{murdoch2019}.

A different perspective about interpretability is given in \citet{molnar2020}, which states that ``Interpretable Machine Learning refers to methods and models that make the behavior and predictions of machine learning systems understandable to humans.'' This definition combines the human-centric aspect of interpretability \citep{doshi2017} and the importance of understanding the cause and effect within a model \citep{thampi2022}, which allows it to inherit all the merits of the definitions that have been discussed so far. Nonetheless, these definitions generally lack a mathematical framework necessary for systematic application.

An explicit way to mathematically incorporate interpretability as part of a supervised learning problem (adapted from \citealp{rudin2022}) is to frame it as the constrained optimization problem
\begin{equation}
\label{eq:rudin_interpret}
\begin{gathered}
\min_{f \in \mathcal{F}} \ \frac{1}{n} \sum_{i=1}^n l(y_i, f(\boldsymbol{x}_i)) + C \cdot \text{InterpretabilityPenalty}(f) \\
\text{subject to InterpretabilityConstraint}(f)
\end{gathered}
\end{equation}
where $n$ is the number of training observations; $f$ is a model chosen from a function class $\mathcal{F}$; $\boldsymbol{x}_i$ is the feature vector of the $i$-th sample and $y_i$ is the corresponding observed response; $l(\cdot, \cdot)$ denote the loss function representing the goodness-of-fit of the model $f$; and $C$ is a hyperparameter representing the trade-off between accuracy and interpretability.

This framework recognizes that the degree of model interpretability should be determined based on the context to which the model is applied. For example, the interpretability criteria for an image recognition model, which processes pixels, will differ from those of an insurance pricing model that mainly uses tabular data. Examples of interpretability constraints include sparsity, monotonicity, decomposability, etc., \citep{rudin2022}. Using this definition facilitates the creation of a common standard for interpretable models in different industries, where domain experts can define a set of interpretability constraints, and a model is deemed interpretable only if it satisfies all these predefined requirements. This approach thus mitigates the issue of subjective interpretation, where different people can disagree about the degree of interpretability that a model has.

%\subsection{Methods to Attain Interpretability} 
\subsubsection{Posthoc Explainability vs Inherent Interpretability}\label{subsec:inter_method}

There are two families of methods that are often adopted to make machine learning models interpretable: \textit{post-hoc explainability} and \textit{inherent interpretability} \citep{harris2024}. Traditionally, the terms \textit{interpretability} and \textit{explainability} are used interchangeably, however, as more research is conducted in developing interpretable machine learning models, it has been pointed out that having a clear distinction between these two terms and approaches is crucial since they serve different purposes and have different implications \citep{rudin2019}.

While there are no formal definitions of \textit{post-hoc explainability} and \textit{inherent interpretability}, there are various aspects that can be used to distinguish these two terms. \citet{murdoch2019} distinguishes model-based (or inherent) interpretability and post-hoc explainability by referring them to two different stages of a data-science life cycle. Inherent interpretability is related to the process of constructing the models, while post-hoc methods are adopted after a model has been built to explain or interpret its prediction. \citet{rudin2022} draws a clear distinction between the two approaches by pointing out that they belong to two different fields of research. Post-hoc methods belong to the field of Explainable AI, where the focus is on developing and applying techniques to approximate or explain the decision of black-box models after they have been trained. This is in contrast to inherent interpretability, which belongs to the Interpretable Machine Learning and where the focus is on developing models that are not black boxes.\footnote{\citet{rudin2019} defines a black box model as either (1) a function that is too complicated for any human to comprehend or (2) a function that is proprietary.}

Post-hoc methods can be classified into different categories depending on what and how the information from the black box models can be extracted. \citet{soa2021} divides these methods into 3 groups:
\begin{enumerate}
    \item Methods to understand relationships between model inputs and outputs (main effects) e.g.\ Partial Dependence Plot \citep{friedman2001}, Individual Conditional Expectation \citep{goldstein2015}.
    \item Methods to identify and visualize interaction effects e.g.\ H-Statistics \citep{friedman2008}, SHAP \citep{lundberg2017}.
    \item Feature importance e.g.\ Permutation Feature Importance \citep{breiman2001}.
\end{enumerate}

In addition to these groups, a common post-hoc approach uses surrogate models \citep{ribeiro2016}—such as decision trees or linear models—to approximate a black-box model's behavior on a local or global scale. These surrogates are trained using the black-box predictions as targets and the original features as inputs.

While post-hoc approaches can provide valuable insight into the inner workings of black-box models, a key limitation is that they only provide an approximation of the logic of black-box models, and these approximations may produce explanations that are unfaithful to the reasoning of the original model, which can result in misleading interpretations \citep{rudin2019}. Another disadvantage of post-hoc methods is that they are vulnerable to adversarial attack, where there are methods to train and manipulate a model such that its underlying logic and learned relationship can't be captured by post-hoc techniques \citep{xin2024}. Due to these shortcomings, it is suggested that in high-stakes domains, instead of using black box models accompanied by explainable machine learning techniques, inherently interpretable models should be the preferred choice. Although there is no consensus on what interpretability means in the machine learning context, there are various models that are widely considered interpretable in actuarial applications, such as linear models, standard decision trees, or generalized additive models. 

Note also that in machine learning it is commonly believed that enhancing predictive performance comes at the expense of interpretability—implying that black-box models yield more accurate predictions than interpretable ones. However, \citet{rudin2019} argues that with well-processed data and meaningful features, interpretable models can perform comparably to black boxes. Moreover, their transparency and ease of troubleshooting enable iterative improvements that may further boost accuracy. An example of this is \citet{wang2023}, who compares the accuracy between interpretable and black-box methods in criminal recidivism prediction, and a key finding is that interpretable models can perform just as well as their uninterpretable counterparts.

\subsection{Our Interpretability Framework} \label{subsec:interpretability}

\subsubsection{Interpretability Definition}

Though there is no consensus on how interpretability can be defined, we can leverage different aspects of this concept that are commonly discussed in the literature to establish a definition of interpretable models in insurance pricing. Using this approach, we combine the interpretations of interpretability in \citet{thampi2022} and \citet{rudin2022} and propose the following definition:

\begin{quote}
An interpretable pricing model follows a set of \textbf{domain-specific} requirements to make the \textbf{cause and effect} within the model \textbf{transparent and easily understood by humans}.
\end{quote}

The emphasis on cause and effect requires a model to offer transparency in its internal logic, specifying how it arrives at a particular prediction, as well as the relationships it has learned from the data. In other words, an interpretable pricing model should give an exact description of how each rating factor influences the output and how the rating factors interact with one another. These attributes are essential in ensuring that the reasoning behind each decision is accessible and understandable to humans.

It is important to point out that under this definition, complex architectures such as neural networks can be incorporated into an interpretable model, and there is no requirement to understand the role and meaning of all model parameters. As long as our model can satisfy all predefined requirements, which have been chosen to ensure the transparency and understandability of the cause and effect, then we can conclude that an interpretable model has been attained.

\subsubsection{Mathematical Formulation}

% \subsection{Problem Specification and Notation} \label{subsec:notation}

The following notations are adopted in this paper unless otherwise specified:
\begin{itemize}
    \item $Y$: the random variable representing the response, which typically represents claim severity or frequency in general insurance pricing $(Y \in \mathbb{R}^+)$. $Y_i$ denotes the response of observation $i$.
    \item $y_i$: the observed response of observation $i$.
    \item $\mathcal{X}$: the feature space, denoting the set of all possible combinations of rating factors.
    \item $\boldsymbol{X} = [X_1, X_2,\dotsc, X_p]^\top$: the $p$-dimensional vector of rating factors $(\boldsymbol{X} \in \mathcal{X}, p \in \mathbb{N})$. The rating factors can also be referred to as covariates, features, or input variables.
    \item $\boldsymbol{X}_i = [X_{i1}, X_{i2},\dotsc, X_{ip}]^\top$: the vector of rating factors associated with the $i$-th observation.
    \item $\phi(Y|\boldsymbol{X})$: the conditional probability density function of the response $Y$ given the rating factors $\boldsymbol{X}$.
    \item $\mu = E(Y|\boldsymbol{X})$: the conditional mean of the response given the input $\boldsymbol{X}$.
\end{itemize}

It is generally assumed in insurance pricing that there exists a function $f(\boldsymbol{X})$, describing the dependence of the response variable on the rating factors ($f: \mathcal{X} \rightarrow \mathbb{R}^+$), such that $\phi(Y|\boldsymbol{X}) = \phi(Y|f(\boldsymbol{X}), \boldsymbol{\sigma})$ where $\boldsymbol{\sigma}$ denote a vector of auxiliary distributional parameters. 

The goal of this research project is to use a parametric function $\hat{f}(\boldsymbol{X}, \boldsymbol{\theta})$ to approximate $f(\boldsymbol{X})$, where $\boldsymbol{\theta} = [\theta_1, \theta_2,\dotsc, \theta_m]^\top$ denote the $m$-dimensional parameter vector $(m \in \mathbb{N})$. The function $\hat{f}(\boldsymbol{X}, \boldsymbol{\theta})$ is represented by a deep learning architecture, and our main interest is to obtain an interpretable representation of this function by imposing various constraints on its functional form and parameters $\boldsymbol{\theta}$.

%% PL: above this was previously a 'notation' section

To formulate the given definition in a mathematical context, we develop a framework outlining how an interpretable model can be constructed in supervised learning, which is a frequently encountered modeling context in insurance pricing. Under this framework, an interpretable pricing model can be expressed as
\begin{equation}
    \label{eq:interpretable_model}
    \hat{f} = \underset{f \in \mathcal{F}}{\mathrm{argmin}} \left(\frac{1}{n} \sum_{i = 1}^n \mathcal{L}(f(\boldsymbol{X}_i, \boldsymbol{\theta}), Y_i) + \sum_{j = 1}^z \omega_j\mathcal{P}_j(\boldsymbol{\theta})\right)
\end{equation}
where
\begin{itemize}
    \item $\mathcal{F}$ denotes a family of functions subject to a predefined functional form,
    \item $\mathcal{L}(\cdot, \cdot)$ is the loss function measuring the discrepancy between the model's prediction and the observed output,
    \item $\mathcal{P}_j(\boldsymbol{\theta})$ denotes the $j$-th interpretability penalty associated with a given interpretability requirement imposed on the model's parameters, and
    \item $\boldsymbol{\omega} = [\omega_1, \omega_2,\dotsc, \omega_z]^\top$ is a vector of penalty parameters representing the trade-off between interpretability and accuracy and can be tuned as part of the training process.    
\end{itemize}

This expression builds upon the mathematical formulation given in \citet{rudin2022}. Training an interpretable model under our framework involves estimating the parameters $\boldsymbol{\theta}$ to minimize an objective function with penalty terms, where the model is constrained to a predefined family of functions $\mathcal{F}$. This indicates that the interpretability requirements can be incorporated by either restricting the functional form of the model or adding penalty terms to the objective function. The use of penalty terms also allows us to enforce other practical requirements other than interpretability.

\subsubsection{Selection of Interpretability Requirements} \label{subsubsec:sel_constraints}
Since our definition aligns with the domain-specific notion of interpretability, a critical task is to specify the set of interpretability requirements specific to general insurance pricing so that the cause and effect within the resulting model are transparent and easily understood by humans. We also need to identify the essential practical requirements that a pricing model should have to ensure the applicability of our modelling framework. Inspecting the classical pricing models, which are widely considered interpretable,  together with the relevant literature in interpretable machine learning, six interpretability and practical requirements are adopted in our interpretability framework including:
\begin{enumerate}
    \item \textbf{Transparency of main effects}: the model should provide an exact description of the univariate relationship between each covariate and the model output.
    \item \textbf{Transparency of interaction effects}: the model should clearly describe which interactions among the input covariates are used and how they influence the final prediction.
    \item \textbf{Quantification of variable importance}: the model should be able to quantify and rank the covariates based on their significance in influencing the model output.
    \item \textbf{Sparsity}: the model should only include important features that are known or perceived to have a significant impact on the final prediction.
    \item \textbf{Monotonicity}: the model should have the ability to capture the known monotonic relationships between certain covariates and the target variable.
    \item \textbf{Smoothness}: the model's prediction should vary in a smooth manner with respect to changes in certain rating factors.
\end{enumerate}

The first three attributes are inherent interpretability requirements in classical pricing models like GLMs and GAMs and are the main features contributing to their widespread acceptance as interpretable models. It is the lack of these features that makes novel machine learning methods such as neural networks often regarded as black boxes. The inclusion of sparsity is intuitive as a sparser model is easier to understand and explain. Models like GLMs or decision trees may cease to be interpretable when there is a large number of variables involved \citep{kuo2020}. Having a sparse model also helps reduce computation costs when complex architecture like neural networks is adopted.

Smoothness and monotonicity are important practical requirements in insurance pricing. They are included to ensure that the relationships captured by an interpretable model are reasonable for actuaries to explain and justify the model's results to stakeholders. Predictions that vary roughly with changes in rating factors can indicate problems with data credibility \citep{richman2024b}. The use of smoothing methods is common in actuarial practice, with \citet{cas2007} utilizing these techniques in general insurance pricing with GLM or \citet{soa2018} discussing the graduation of mortality rates. In terms of monotonicity, certain variables used in general insurance pricing are expected to have monotonic relationships with claim costs (e.g.\ worsening bonus-malus score should generally lead to a higher premium). Enforcing monotonicity with these variables can thus enhance the model's interpretability and acceptance by stakeholders.

\section{Modeling Framework} \label{sec:mf}
The modeling approach developed in this paper builds on the Neural Additive Model (NAM) proposed by \citet{agarwal2021}, with targeted extensions to account for pairwise interaction effects and incorporate essential interpretability requirements in insurance pricing. Specifically, we implement architectural constraints to enforce monotonicity, introduce penalty terms in the optimization algorithm to promote smoothness, and design a three-stage process for effective variable selection. These enhancements ensure that Actuarial NAM aligns with the established interpretability framework while preserving the predictive power of deep learning.

\subsection{Model Architecture} \label{subsec:model_architecture}

With Actuarial NAM, each input feature or pairwise interaction is assigned a dedicated neural network, referred to as a subnetwork, to independently learn its relationship with the model's output. Our model contains a total of $m_1 + m_2$ subnetworks with $m_1$ and $m_2$ respectively allocated for main and pairwise interaction effects. 

Each subnetwork consists of multiple layers of interconnected neurons, with the output layer producing a single value representing the subnetwork's contribution to the overall prediction. A continuous variable is assigned a single neuron in the input layer of its corresponding subnetworks while categorical variables are represented using one-hot encoding where a feature with $k$ levels is transformed into a sparse $k$-dimensional vector with a single element set to 1 and the remaining elements set to 0. The one-hot encoded vector for a categorical feature $X_i$ with $k$ categories ($a_1, a_2, \ldots, a_k$) can be expressed as
$
    \bigl[\mathbb{I}_{\{X_i = a_1\}}, \mathbb{I}_{\{X_i = a_2\}},\dotsc, \mathbb{I}_{\{X_i = a_k\}}\bigr]^\top.
$

Let $L_i$ be number of layers and $q^{[i]}_l$ be the number of neurons in the $l$-th layer of the $i$-th subnetwork, where $1 \leq i \leq m_1 + m_2$ and $1 \leq l \leq L_i$. We have $q_{L_i}^{[i]} = 1$ for all subnetworks while the size of the input layer $q_{1}^{[i]}$ will depend on the type of variable (continuous or categorical) and the effect (main or interaction) that each subnetwork is designed to model. As each subnetwork is a feedforward neural network, the $l$-th layer can be defined as

\begin{equation}
    \boldsymbol{a}^{(l)}_i : \mathbb{R}^{q_{l-1}^{[i]}}\ \rightarrow \mathbb{R}^{q_{l}^{[i]}}, \quad \boldsymbol{a}^{(l)}_i(\boldsymbol{v}) = \left[a_{i1}^{(l)}(\boldsymbol{v}),a_{i2}^{(l)}(\boldsymbol{v}),\dotsc, a_{iq_l^{[i]}}^{(l)}(\boldsymbol{v}) \right], \text{ with }
\end{equation}
\begin{equation}
a_{ij}^{(l)}(\boldsymbol{v}) = \psi^{(l)}_i \left(b_{ij}^{(l)} + \boldsymbol{w}_{ij}^{(l)\top}\boldsymbol{v} \right); \hspace{1em} j = 1, \ldots, q_l^{[i]} \text{ and } 2 \leq l \leq L_i
\end{equation}
where $\psi^{(l)}_i(\cdot)$ is the activation function, $b_{ij}^{(l)}$ and $\boldsymbol{w}_{ij}^{(l)}$ are the bias and weights, respectively, associated with the $j$-th neuron in the $l$-th layer of the $i$-th subnetwork.

\begin{figure}[h]
    \centering
    
    \centering
    \resizebox{\linewidth}{!}{\input{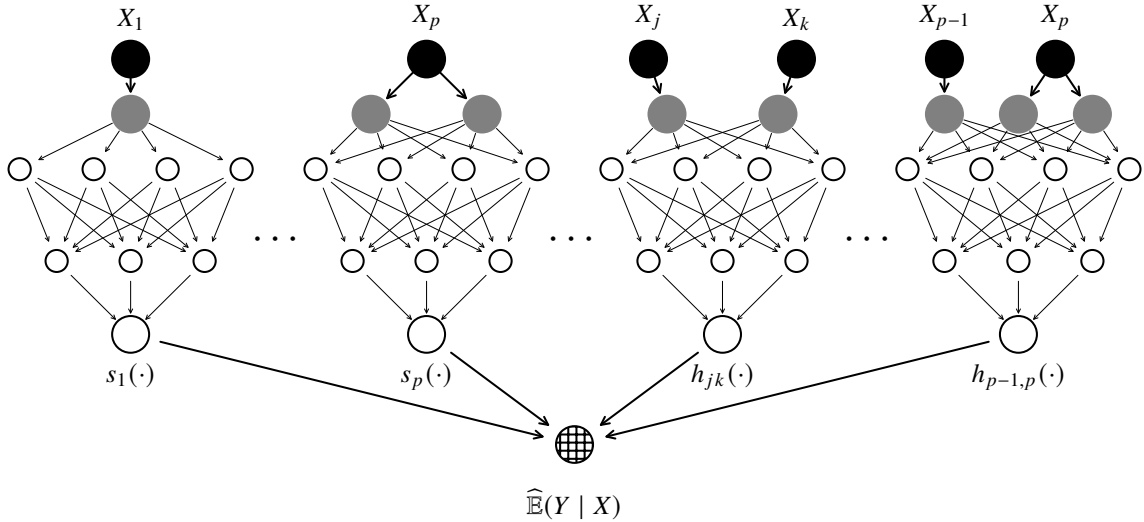}}

    \caption{An illustration of the the proposed model's architecture. Black circles represent raw values of the features, grey circles represent the preprocessed versions (the one-hot encoded versions of categorical features, e.g. $X_p$ is binary categorical here), white circles represent the hidden neurons, and the cross-hatched circle represents the output neuron. %The number of hidden neurons and hidden layers shown here is just for illustration purposes.
    }
    \label{img:model}
\end{figure}

The neural additive model adopted in this paper can be expressed as:
\begin{equation}
    g(\mu) = \beta + \boldsymbol{\alpha}_{1}\sum_{i = 1}^ps_i(X_i) + \boldsymbol{\alpha}_{2}\sum_{j = 1}^p\sum_{\substack{k=1 \\ k \ne j}}^j h_{jk}(X_j, X_k)
\end{equation}
where $g(\cdot)$, $\beta$, $\boldsymbol{\alpha}_{1}$, and $\boldsymbol{\alpha}_{2}$ respectively denote the monotone link function, the bias term, and vector of weights connecting the output of the univariate and bivariate subnetworks with the output neuron of the models; $s_i(X_i)$ and $h_{jk}(X_j,X_k)$ respectively represent the shape functions of univariate and bivariate effects learned by the subnetworks. A graphical visualization of our proposed model is provided in \cref{img:model}.

% \begin{figure}[h]
%     \centering
%     \includegraphics[width=0.75\textwidth]{images/Model.png}
%     \caption{The architecture of the proposed model}
%     \label{img:model}
% \end{figure}

Under this modeling framework, the transparency of main and interaction effects is embedded in the model architecture as each subnetwork is dedicated to a single or pair of variables, and an exact description of the relationships between the input covariate and output can be obtained through visualization of the charts or heatmaps that summarize each shape function. This gives us a global interpretation of the model's internal logic and is also the main reason why only main and pairwise interaction terms are considered in our modeling framework as it is challenging to obtain the visualization for effects of higher order. Additionally, the output of Actuarial NAM can be locally explained since the additive decomposition adopted by the model facilitates the interpretation of how different components of the input contribute to the final prediction. A quantification of the importance of each main and pairwise term can also be obtained by evaluating the variance of their corresponding shape function. This measure can be adopted to create a sparse model. Furthermore, the use of parallel subnetworks facilitates the implementation of smoothness and monotonicity requirements as they can be imposed directly on the architecture of the relevant subnetwork, independent of other model components. 

Besides interpretability, Actuarial NAM (or ANAM) offers various modeling benefits, which justify our decision to embed neural networks in a GAM framework compared to splines or decision trees. Firstly, ANAM can support multi-task learning (e.g.\ model claim frequency and severity simultaneously), and also facilitate flexible probabilistic distributions, which are essential in actuarial modeling, by simply increasing the number of output neurons in the last layer. This setup (see, for example, \ref{append_B}) allows shared learning signals across related tasks, improving both the generalization and predictive capability of the model. Secondly, because ANAM is based on neural networks, it can handle diverse data types, such as text and images, which traditional methods like splines or decision trees manage inefficiently. By assigning dedicated subnetworks, we can effectively process textual or image information and understand how they contribute to the final prediction, independent of other input features. Thirdly, ANAM can be extended to incorporate advanced modeling techniques, which enhance its flexibility and performance. For example, methods like entity embedding can be easily incorporated into our modeling framework, allowing ANAM to effectively handle high-cardinality categorical features commonly found in actuarial datasets, such as occupations, postal codes, or vehicle models.

\subsection{Monotonicity} \label{subsec:monotonic}

Under the chosen interpretability framework, to ensure that the output of the model is monotonic with respect to certain input variables, monotonicity requirement can be incorporated by adding a penalty term in the objective function or by imposing architectural restrictions on the shape functions of the relevant variables. The use of monotonicity penalties with NAM or artificial neural networks has been discussed in \citet{chen2022} and \citet{richman2024b}. However, this approach doesn't guarantee that the resulting model will strictly produce monotonic relationships, as there are other factors influencing the objective function in the optimization process besides the penalty terms for monotonicity. 

In contrast, our approach is to adopt a method known as \textit{monotonic lattice regression} \citep{garcia2009, gupta2016}, instead of an FNN, to model the shape functions containing variables that are expected monotonic relationship with the model output. Since Actuarial NAM only models main and pairwise interaction effects, the mathematical formulation of monotonic lattice regression with two-dimensional input will be provided. The one-dimensional scenario can be easily derived from our formulation.

A lattice is a grid of numbers with the value at each vertex representing the function value at a given input. Let $M_i \in \mathbb{N}$ be the number of discretization points along the $i$-th input dimension. Then, a lattice with two-dimensional input $(X_j,X_k)$ is a grid of $M_j \times M_k$ parameters placed at natural numbers that span the rectangle $[0, \, M_j - 1] \times [0, \, M_k - 1]$ in a Euclidean space. Inputs in the $i$-th dimension are required to be pre-processed and scaled to the range $[0, \, M_i - 1]$. For categorical features, $M_i$ represents the number of unique categories, while for continuous variables, the following formula is adopted for scaling:
\begin{equation}
    \label{eq:min_max}
    X_i' = (M_i - 1) \times \frac{X_i - X_i^{\text{min}}}{X_i^{\text{max}} - X_i^{\text{min}}}
\end{equation}
where $ X_i^{\text{min}}$ and $X_i^{\text{max}}$ are the minimum and maximum values of the input $X_i$.

 The lattice regression model with two-dimensional input $(X_j,X_k)$ is represented by the function $m: \mathbb{R}^2 \rightarrow \mathbb{R}$, defined using bilinear interpolation:
\begin{equation}
    \label{eq:lattice}
    m(X_j,X_k) = (1 - \gamma_1)(1 - \gamma_2) \lambda_{a b} + \gamma_1(1 - \gamma_2) \lambda_{(a+1) b} + (1 - \gamma_1)\gamma_2 \lambda_{a (b+1)} + \gamma_1\gamma_2 \lambda_{(a+1)(b+1)},
\end{equation}
where 
\begin{itemize}
    \item $\lambda_{ab}$ denote the lattice value at the vertex corresponding to the $a$-th level of $X_j$ and $b$-th level of $X_k$ ($a,b \in \mathbb{N}$; $0 \leq a \leq M_j - 2$; $0 \leq b \leq M_k - 2$),
    \item $\gamma_1 = X_j' - a$,
    \item $\gamma_2 = X_k' - b$,
    \item $X_j'$ and $X_k'$ are calculated as in \cref{eq:min_max}.
\end{itemize}
To ensure that the model output is monotonic to the input of a given dimension, monotonicity requirements are imposed on the lattice parameters $\lambda_{a b}$. For example, if the model is expected to be monotonically increasing along the $j$-th dimension, then the following constraints are enforced:
\begin{equation}
    \label{eq:mono_constraints} 
    \lambda_{(a+1) b} - \lambda_{a b} \geq 0, \quad \text{for all } a \in \{0,\, 1, \ldots, M_j - 2\}, \, b \in \{0,\, 1, \ldots, M_k - 1\}.
\end{equation}

From the above mathematical formulation, it's easily seen that in case the input is one-dimensional, the lattice regression model reduces to a linear spline with vertices playing the role of knots, connecting the segments of the piecewise linear function.

Lattice regression is chosen to model monotonic and partially monotonic shape functions in our Actuarial NAM framework for four reasons. Firstly, with constraints being applied directly to the model parameters and an appropriate training algorithm, this method can produce outputs that are guaranteed to be monotonic, ensuring the reliability of the model \citep{gupta2016}. Secondly, unlike the shape functions produced by artificial neural networks where interpretability is obtained via visualization, lattice regression produces interpretability at the parameter level, offering greater insight into the model's behavior. Thirdly, this method can accommodate different shape constraints (e.g.\ convex, smooth), providing the flexibility to model a wide range of relationships inherent in actuarial data. Finally, lattice regression can be seamlessly trained using gradient descent, enabling straightforward integration with neural network training processes.

\subsection{Smoothness} \label{subsec:smoothness}
Smoothness requirement will be facilitated via a penalty term added to the objective function. The use of roughness penalties to ensure a smooth relationship between the input covariates and model output is a common approach in machine learning and actuarial modeling \citep{eilers1996, richman2024b}. In our modeling framework, the roughness penalties will only be applied to the shape functions of the main effects to avoid having an excessive number of penalty terms that could potentially degrade the model's predictive performance. The smoothness of a function with respect to an input feature $X_i$ can be quantified using the second derivative so the roughness penalty can be expressed as:
\begin{equation}
    \label{eq:smooth_loss}
    \mathcal{P}_{\text{smooth}} = \omega_{\text{smooth}} \sum_{s \in \mathcal{S}} \int_{-\infty}^{\infty} \left| \frac{\partial^2 s}{\partial x_i^2} \right| \mathrm{d} x_i
\end{equation}
where $\mathcal{S}$ is the set of shape functions where smoothness is required, and $\omega_{\text{smooth}}$ is the hyperparameter representing the trade-off between prediction accuracy and the smoothness of the functions in $\mathcal{S}$.

In practice, integrating a shape function produced by a neural network across its entire domain is computationally challenging. Therefore, we approximate the penalty term in \cref{eq:smooth_loss} using finite differences, a method commonly employed in graduation techniques within actuarial science \citep{whittaker1922, henderson1924}.
\begin{equation}
    \label{eq:smooth_loss_approx}
    \mathcal{P}_{\text{smooth}} \approx \omega_{\text{smooth}} \sum_{s \in \mathcal{S}}\sum_{i=2}^{n-1} \left| \frac{s(x_{i+1}) - 2s(x_i) + s(x_{i-1})}{h^2} \right|
\end{equation}
where \(x_1 < x_2 < \dotsb < x_n\) are sampled from the domain of a shape function \(s\) with uniform spacing \(h = x_{i+1} - x_i\).

\subsection{Sparsity} \label{subsec:sparsity}
Since each covariate and pairwise interaction term is assigned a dedicated subnetwork or lattice model, the Actuarial NAM framework entails a large number of parameters, which grows rapidly with an increase in the size of input features. Consequently, computational cost becomes a significant concern, making common methods like adding regularization terms in the objective function \citep{hastie1986, ravikumar2009, xu2023} or using forward/backward stepwise selection less desirable. Our modeling framework enforces sparsity requirement by following a three-stage process to select significant terms. Although multiple NAMs will be fitted in this variable selection process, the computational cost is controlled by limiting the size of each subnetwork and the number of covariates and interaction terms that each NAM has to consider.
\begin{enumerate}
    \item \textbf{Main Effect Selection}: Train an ensemble of NAMs with only main effects using all input covariates. The rationale for using an ensemble rather than a single model is that neural networks can produce outputs that vary widely with different initializations of model parameters \citep{richman2022}. Using an ensemble helps ensure a stable output for the variable selection process. The shape functions are modeled using either a shallow neural network or a lattice model with few vertices to limit computational cost. The importance of each variable $i$ is measured by calculating the average variance of its corresponding shape function $\frac{1}{N}\sum_{k=1}^N\frac{1}{n-1}\sum_{j=1}^n\bigl(s_i^{(k)}(x_i^{(j)})\bigr)^2$, where $n$ is the number of training instances, $N$ is the number of models in the ensemble, $s_i^{(k)}(\cdot)$ is the shape function of the $i$-th variable in the $k$-model of the ensemble, and $x_i^{(j)}$ is the value of the $i$-th variable in the $j$-th training instance. Select the top $K_1$ variables with the highest score based on this measure.
    \item \textbf{Interaction Effect Selection}: Based on the hierarchy principle \citep{bien2013, james2023}, we only consider a pairwise term if both of its associated main effects have been selected in the first stage. This is often referred to as the \textit{strong heredity constraint} \citep{yang2021}, adopted to limit computational cost. Starting with the model containing only the significant main effects obtained from step 1, we record its validation loss as a baseline. Each candidate pairwise interaction term is then added individually to this model, resulting in $\frac{K_1(K_1-1)}{2}$ models to consider, each with all significant main effects and a single pairwise term. Importantly, when adding a pairwise term to this baseline model, only the parameters of the subnetwork or lattice model corresponding to the pairwise term and its associated main effects are trained, while all other parameters are kept unchanged. This approach further reduces the computational cost. For each model, we evaluate the change in validation loss compared to the baseline, selecting the top $K_2$ interaction terms that yield the greatest reduction in validation loss for inclusion in the final model.
    \item \textbf{Fine tuning}: All selected main and interaction effects are re-trained simultaneously as a full model.
\end{enumerate}

In the above procedure, only in step 3 are penalty terms included in the objective function to enforce all relevant interpretability and practical requirements. In the first two stages, NAMs are trained solely by minimizing the negative log-likelihood. Additionally, we do not use a data-driven approach to determine the values of $K_1$
and $K_2$. Instead, this is left to qualitative judgment considering domain knowledge, data features, and the desired sparsity level, as the optimal number of terms varies depending on the context and user preference. Even if an additional term may improve validation loss, the improvement might not be significant enough to justify the added computational cost, making this decision context-dependent.

\subsection{Identifiability}
Identifiability is not directly related to interpretability. However, since our model is constructed under a GAM framework, it is crucial to impose certain constraints to separate the effects of each component, ensuring that each shape function represents a unique contribution to the model. To ensure that the model is identifiable, the following constraints are enforced:
\begin{equation}
\label{eq:identifiability}
\begin{aligned}
    &\int_{-\infty}^{\infty} s_i(x_i) \, \mathrm{d} F_i(x_i) = 0, \\
    &\iint_{\mathbb{R}^2} h_{ij}(x_i, x_j) \, \mathrm{d} F_{ij}(x_i, x_j) = 0, \\
    &\iint_{\mathbb{R}^2} s_i(x_i) \, h_{ij}(x_i, x_j) \, \mathrm{d}F_{ij}(x_i, x_j) = 0,
\end{aligned}
\quad \forall \, i, j \in \{1, 2, \dots, p\}, \, i \ne j,
\end{equation}
where $F(\cdot)$ denote the corresponding cumulative distribution function.

The first two constraints are adopted to avoid mixing the relationship between each shape function and the bias term, ensuring that each shape function is centered and its effects are distinct from any constant offset. The third constraint, often known as \textit{marginal clarity constraint}, is adopted to make each main effect and its associated pairwise interaction terms orthogonal, thus avoiding any confounding between main and interaction effects \citep{yang2021}.

In practice, performing the integration is computationally challenging, especially when working with non-standard distributions. To simplify implementation, the first two constraints are enforced by centering each shape function using its sample means:
\begin{equation}
\begin{aligned}
    s_i(x_i) &\gets s_i(x_i) - \frac{1}{n} \sum_{k=1}^n s_i(x_i^{(k)}), \\
    h_{ij}(x_i, x_j) &\gets h_{ij}(x_i, x_j) - \frac{1}{n} \sum_{k=1}^n h_{ij}(x_i^{(k)}, x_j^{(k)}),
\end{aligned}
\end{equation}
where $n$ is the number of training instances and $x_i^{(k)}$ denotes the value of the $i$-th feature for the $k$-th training instance.

With the marginal clarity constraint, we approximate the degree of orthogonality between each main effect $s_i(x_i)$ and its associated interaction terms $h_{ij}(x_i,x_j)$ using an empirical measure. This measure is incorporated as a penalty term in the objective function, which can be written as:
\begin{equation}
    \label{eq:mc_penalty}
    \mathcal{P}_{\text{mc}} = \omega_{\text{mc}} \sum_{i = 1}^p\sum_{\substack{j=1 \\ j \ne i}}^i \left| \frac{1}{n} \sum_{k=1}^n s_i(x_i^{(k)}) h_{ij}(x_i^{(k)}, x_j^{(k)}) \right|
\end{equation}
where $\omega_{\text{mc}}$ is the hyperparameter representing the trade-off between marginal clarity and other optimization goals; $n$ is the number of training instances; and $x_i^{(k)}$ denotes the value of the $i$-th feature for the $k$-th training instance.

In the ideal scenario, we would want $P_{\text{mc}}$ to be exactly zero, indicating perfect orthogonality. However, due to the presence of multiple objectives within the objective function, a small deviation from zero is acceptable in practice.

\subsection{Optimization Algorithm}

The Actuarial NAM is trained by optimizing model parameters to minimize the negative log-likelihood of the response variable, with added penalties for smoothness and marginal clarity, while adhering to monotonicity requirements. Let $\mathcal{M}$ denote the set of all monotonic and partially monotonic shape functions modeled using lattice regression, with $\boldsymbol{\lambda}_i \in \mathbb{R}^{z_i \times 1}$ representing the parameter vector for the $i$-th function in $\mathcal{M}$ ($1 \leq i \leq |\mathcal{M}|$), where $z_i$ is the number of lattice parameters for function $i$. The monotonicity constraints for each $\boldsymbol{\lambda}_i$ are represented by a matrix $A_i \in \mathbb{R}^{v_i \times z_i}$, where $v_i$ is the number of monotonicity constraints for the $i$-th function. With these notations, the set of monotonicity constraints for the function $i$ in $\mathcal{M}$ can be expressed as $A_i\boldsymbol{\lambda}_i \leq \boldsymbol{0}$, with each row of the sparse matrix $A_i$ containing a single 1 and -1, while all other entries are 0. This setup enforces a system of linear inequalities that ensures each function in $\mathcal{M}$ adheres to the required monotonic behavior.

Let $\boldsymbol{\lambda} = [\boldsymbol{\lambda}_1, \boldsymbol{\lambda}_2,\dotsc, \boldsymbol{\lambda}_{|\mathcal{M}|}]^\top$ denote the vector of all parameters of the shape functions in $\mathcal{M}$ and $\boldsymbol{\tau}$ denote the vector of all remaining model parameters. The objective function of our modeling framework can be expressed as:
\begin{equation}
\label{eq:objective_function}
    \mathcal{C}(\boldsymbol{Y}, \boldsymbol{\theta}) = \sum_{i = 1}^n \log\left( \phi(y_i|\hat{f}(\boldsymbol{x}_i, \boldsymbol{\theta})) \right) + \mathcal{P}_{\text{smooth}} + \mathcal{P}_{\text{mc}}
\end{equation}
subject to
\begin{equation}
    A_j\boldsymbol{\lambda}_j \leq \boldsymbol{0}, \quad \forall \, j \in \{1,\dotsc,|\mathcal{M}|\}
\end{equation}
where $n$ is the number of training instances; $\boldsymbol{\theta}=[\boldsymbol{\lambda},\boldsymbol{\tau}]^\top$; $\hat{f}$ is the function produced by the Actuarial NAM; and $\mathcal{P}_{\text{smooth}}$ and $\mathcal{P}_{\text{mc}}$ are defined in \cref{eq:smooth_loss_approx,eq:mc_penalty} respectively.

To obtain an estimation $\hat{\boldsymbol{\theta}}$ of $\boldsymbol{\theta}$ that minimizes the objective function $\mathcal{C(\boldsymbol{Y}, \boldsymbol{\theta})}$, we adopt the projected gradient descent approach that involves two main steps in each iteration. First, all model parameters $\hat{\boldsymbol{\theta}}$ are updated in the direction of the steepest descent. Following this gradient descent update, a projection onto the feasible set, represented by the linear inequalities for monotonicity constraints, is applied specifically to the lattice parameters $\hat{\boldsymbol{\lambda}}$ to ensure they satisfy the monotonicity requirements. The algorithm is continued until either the change in the objective function falls below a predefined threshold or a maximum number of iterations is reached.

 Let $\eta$ be the learning rate of the gradient descent algorithm, $\boldsymbol{\theta}^{(t)}=[\boldsymbol{\lambda}^{(t)},\boldsymbol{\tau}^{(t)}]^\top$ be the estimates of the model parameters after iteration $t$, and $\nabla_{\boldsymbol{\theta}} \mathcal{C}(\cdot)$ is the gradient of the objective function with respect to parameters $\boldsymbol{\theta}$. The training procedure for the Actuarial NAM is summarized in \cref{algo:anam_train}.

\begin{algorithm} 
\caption{Training an Actuarial NAM} \label{algo:anam_train}
\begin{algorithmic}[1]
    \STATE \textbf{Input}: Training data $\{\boldsymbol{x}_i, y_i\}_{i=1}^n$, initial parameters $\boldsymbol{\theta}^{(0)} = [\boldsymbol{\lambda}^{(0)}, \boldsymbol{\tau}^{(0)}]^\top$, learning rate $\eta$, maximum iterations $T_{\text{max}}$, convergence threshold $\epsilon$, set of monotonic and partially monotonic shape functions $\mathcal{M}$
    \STATE \textbf{Output}: Estimated parameters $\hat{\boldsymbol{\theta}}$
    \STATE Initialize $t \leftarrow 0$
    \REPEAT
        \STATE Compute gradient $\nabla_{\boldsymbol{\theta}} \mathcal{C}(\boldsymbol{Y}, \boldsymbol{\theta}^{(t)})$
        \STATE Update parameters: $\boldsymbol{\theta}^{(t+1)} \leftarrow \boldsymbol{\theta}^{(t)} - \eta \nabla_{\boldsymbol{\theta}} \mathcal{C}(\boldsymbol{Y}, \boldsymbol{\theta}^{(t)})$
        \FOR{$j = 1$ to $|\mathcal{M}|$}
            \STATE Project $\boldsymbol{\lambda}_j^{(t+1)}$ onto the feasible set defined by $A_j \boldsymbol{\lambda}_j \leq \boldsymbol{0}$ using Dykstra's algorithm (Algorithm 2)
        \ENDFOR
        \STATE $t \leftarrow t + 1$
    \UNTIL{$||\boldsymbol{\theta}^{(t)} - \boldsymbol{\theta}^{(t-1)}|| < \epsilon$ or $t \geq T_{\text{max}}$}
    \STATE \textbf{return} $\hat{\boldsymbol{\theta}} \leftarrow \boldsymbol{\theta}^{(t)}$
\end{algorithmic}
\end{algorithm}

The projection of $\hat{\boldsymbol{\lambda}}_j$ onto the feasible sets defined by  $A_j\boldsymbol{\lambda}_j \leq \boldsymbol{0}$ can be formulated as finding $\boldsymbol{\lambda}_j^*$ in the feasible region defined by the linear inequality constraints that minimize the distance to $\hat{\boldsymbol{\lambda}}_j$. This can be expressed as
\begin{equation}
\boldsymbol{\lambda}_j^* = \arg \min_{\boldsymbol{\lambda}_j} \frac{1}{2} \|\boldsymbol{\lambda}_j - \hat{\boldsymbol{\lambda}}_j\|^2 \quad \text{subject to} \quad A_j \boldsymbol{\lambda}_j \leq \boldsymbol{0}
\end{equation}

We solve this constrained optimization problem by adopting Dykstra's projection algorithm \citep{dykstra1983, boyle1986}, which is a commonly used method with monotonic lattice regression. Let $\mathcal{D}_i$ denote the feasible set represented by the $i$-th constraint in $A_j\boldsymbol{\lambda}_j \leq \boldsymbol{0}$, i.e., 
$\mathcal{D}_i = \{ \boldsymbol{\lambda}_j \mid (A_j)_i\boldsymbol{\lambda}_j \leq 0 \}$ 
where $(A_j)_i$ denotes the $i$-th row of $A_j$. 

The algorithm works by iteratively projecting the estimate of $\boldsymbol{\lambda}_j$ onto each feasible set $\mathcal{D}_i$ until convergence or the maximum number of iterations is attained. Using Dykstra's projection algorithm with a sufficient number of iterations can guarantee that the estimate converges to a solution within the intersection of all feasible sets if it exists \citep{dykstra1983, boyle1986}. We provide a summary of using Dykstra's projection method to enforce monotonicity requirements in our modeling framework in \ref{append_C}.

\section{Data Analysis} \label{sec:da}

\subsection{Hyperparameter Tuning}
The data for model development is split into training, validation, and test sets with the training data used for fitting the model, validation data used for hyperparameter tunings, and test data used for evaluating the final model's predictive performance on unseen observations. In the hyperparameter tuning process, different combinations of hyperparameters will be explored, and the one with the best performance on the validation set will be selected as the final model. The key hyperparameters that can be tuned in our modeling framework include:
\begin{enumerate}
    \item The number of hidden layers for each subnetwork.
    \item The number of neurons in the hidden layers for each subnetwork.
    \item The subnetworks' activation function.
    \item The number of vertices for lattice regression models.
    \item The smoothing parameter $\omega_{\text{smooth}}$ to control the smoothness of the model output.
    \item The regularization strength $\omega_{\text{mc}}$ of marginal clarity constraint.
    \item The learning rate for gradient descent algorithm $\eta$.
\end{enumerate}

It is important to point out that the more hyperparameters are tuned, the larger the number of combinations that need to be evaluated, which can significantly increase computational cost. Therefore, in practice, selecting which hyperparameters to tune depends on the characteristics of the data and the nature of the problem. Focusing on a few key parameters can often yield an efficient balance between model performance and computational efficiency.

\subsection{Model Evaluation} \label{subsec:model_eval}
The models that are implemented and compared with Actuarial NAM in this paper can be divided into three groups.
\begin{enumerate}
    \item \textbf{Classical insurance pricing models}: GLM and GAM.
    \item \textbf{``Interpretable" machine learning models}: LocalGLMnet and Explainable Boosting Machine (EBM), which is a tree-based version of GAM \citep{lou2013}.
    \item \textbf{Black-box machine learning models}: Neural network and Gradient Boosting Machine (GBM).
\end{enumerate}

Three metrics are used to evaluate the predictive performance of the models, capturing both point predictions and distributional forecasts.
\begin{enumerate}
    \item \textbf{Negative Log-Likelihood} (NLL): measure the accuracy of the distributional forecast. NLL is a \textit{proper scoring rule} \citep{gneiting2007}. The lower the NLL, the better the fit of the predicted distribution to the observed data.
    \begin{equation}
        \text{NLL} = - \frac{1}{n^*}\sum_{i=1}^{n^*}\log(\hat{\phi}_i(y_i))
    \end{equation}
    where $\hat{\phi}_i(\cdot)$ is the predicted density function for testing instance $i$ and $n^*$ is the number of testing instances.
    \item \textbf{Root Mean Squared Error} (RMSE): measure the accuracy of point prediction. The lower the RMSE, the better the accuracy of the central estimate. A drawback of this metric is that it is sensitive to outliers due to the use of square differences.
    \begin{equation}
        \text{RMSE} = \sqrt{\frac{1}{n^*}\sum_{i=1}^{n^*}(y_i - \hat{\mu}_i)^2}
    \end{equation}
    where $\hat{\mu}_i$ is the predicted mean of the response of testing instance $i$.
    \item \textbf{Mean Absolute Error} (MAE): measure the accuracy of point prediction. The lower the MAE, the better the predictive performance. MAE uses absolute difference so it is not as sensitive to large discrepancies as RMSE is.
    \begin{equation}
        \text{MAE} = \frac{1}{n^*}\sum_{i=1}^{n^*}|y_i - \hat{\mu}_i|.
    \end{equation}
\end{enumerate}

\subsection{Synthetic Datasets} \label{subsec:syn_data}
\subsubsection{Synthetic Data Generation Mechanism} \label{subsec:syn_data_mecha}
Testing the Actuarial NAM on synthetic datasets enables us to evaluate its ability to recover true underlying relationships and make accurate predictions when the ground truth includes complex, non-linear patterns that the model is designed to capture. For each observation \( i \) in the dataset, a synthetic function first generates the mean \( \mu_i \), which is then used to sample the actual response according to \( Y_i | \boldsymbol{X}_i \sim \text{Gamma}(\mu_i, \phi) \). This approach mirrors the characteristics of claim severity in an insurance portfolio, with \( \phi \) representing the dispersion parameter. The covariates, \( \boldsymbol{X} = [X_1, X_2, \ldots, X_{10}] \), are independently and identically distributed as \( X_i \sim \mathcal{U}(-1, 1) \) for \( i = 1, \dots, 10 \).

In this study, we focus exclusively on modeling the mean response. To assess model performance under different noise conditions, we generate two datasets, each containing 50,000 observations. One dataset simulates a low-noise environment with \( \phi = 1 \), while the other represents a high-noise setting with \( \phi = 5 \). This setup allows us to evaluate the model’s interpretability and prediction accuracy across varying noise intensities, with the assumption that variability is homogeneous across all data points, as indicated by the constant dispersion parameter.

The mean of the response variable is defined as a non-linear function of the covariates, using a log link function, expressed as follows:
\begin{equation} \log(\mu) = \beta + f_1(X_1) + f_2(X_2) + f_{34}(X_3, X_4) + f_{56}(X_5, X_6) + f_{78}(X_7, X_8) \end{equation} 
where $\beta$ is the bias term, and each function $f_i$ or $f_{jk}$ captures specific patterns in the data. The exact functional forms are given as follows:
\begin{equation}
        % \left\{ \begin{array}{l}
        \begin{aligned}
        f_1(X_1) &= |X_1| \sin(8X_1) \\
        f_2(X_2) &= 0.5\sin^3(8X_2) - 0.25\cos(4X_2) + 0.25X_2^2 \\
        f_{34}(X_3, X_4) &= -(X_3 + 0.5) \mathrm{e}^{-X_4} \\
        f_{56}(X_5, X_6) &= 1.5\sin(2\pi (X_5 - 0.5)(X_6 + 0.5)) \\
        f_{78}(X_7, X_8) &= \text{sign}\left( 50 \left( \sin\left(10X_7\right) + 0.5 \right) \right) \times \text{sign}\left( 50 \left( \sin\left(10X_8\right) - 0.5 \right) \right) \\
    % \end{array} \right.
    \end{aligned}
\end{equation}

\begin{figure}[h]
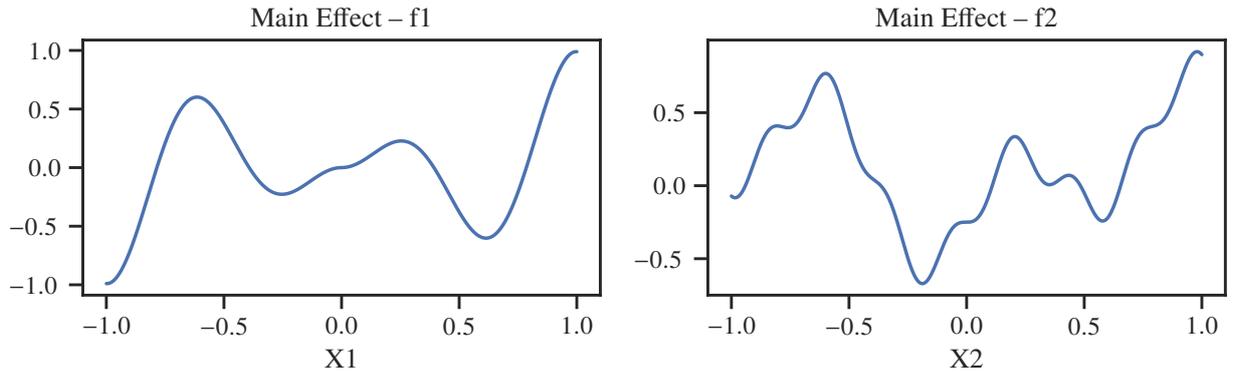

    \centering
    \begin{minipage}{0.5\linewidth}
        \centering
        \resizebox{\linewidth}{!}{\input{images/main_effect_X1.pgf}}
    \end{minipage}%
    \hfill
    \begin{minipage}{0.5\linewidth}
        \centering
        \resizebox{\linewidth}{!}{\input{images/main_effect_X2.pgf}}
    \end{minipage}
    \caption{Ground truth – Main effects}
    \label{img:ground_truth_main}
\end{figure}

\begin{figure}[h]
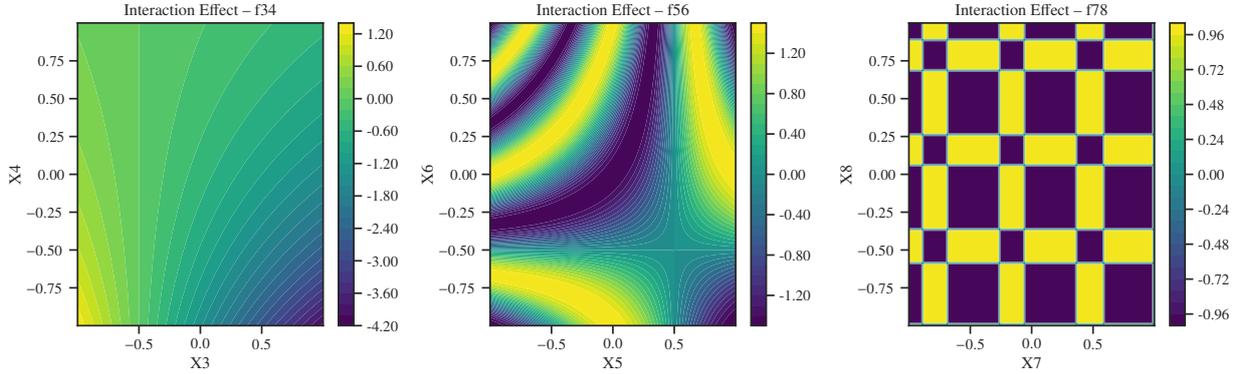

    \centering
    \begin{minipage}{0.33\linewidth}
        \centering
        \resizebox{\linewidth}{!}{\input{images/interaction_f34.pgf}}
    \end{minipage}%
    \hfill
    \begin{minipage}{0.33\linewidth}
        \centering
        \resizebox{\linewidth}{!}{\input{images/interaction_f56.pgf}}
    \end{minipage}
    \begin{minipage}{0.33\linewidth}
        \centering
        \resizebox{\linewidth}{!}{\input{images/interaction_f78.pgf}}
    \end{minipage}
    \vspace{-1em}
    \caption{Ground truth - Interaction Effects}
    \label{img:ground_truth_interaction}
\end{figure}

\Cref{img:ground_truth_main,img:ground_truth_interaction} provide visualizations of the ground truth, illustrating the main and pairwise interaction effects, respectively.

The synthetic data generation function presents several key characteristics that highlight the intended testing scenarios for Actuarial NAM. Firstly, the function has an additive structure with only main and pairwise interaction effects, which are features that Actuarial NAM is specifically designed to capture effectively. Secondly, while we generate ten predictors, $X_1$ through $X_{10}$, only eight of these are actually used in determining the mean response; $X_9$ and $X_{10}$ serve as purely spurious noise inputs. Including these irrelevant predictors allows us to assess the efficiency of our proposed three-stage variable selection process, testing its ability to filter out noise and retain only the meaningful variables. Thirdly, the choice of ground truth functions, as shown in \cref{img:ground_truth_main,img:ground_truth_interaction}, reflects a careful balance between smooth and complex shapes in the main effects. While the simulated main effects are mostly smooth curves, they contain some jaggedness to simulate real-world irregularities. We aim for the Actuarial NAM, with its neural network architecture, to handle these complex shapes while maintaining interpretability in terms of smoothness. Additionally, the three pairwise interaction terms serve distinct purposes in testing the model's capacity for handling interactions of varying complexity. For instance, $f_{34}$ introduces a monotonic decreasing relationship with $X_3$, enabling us to evaluate the efficiency of lattice regression. The heatmap for $f_{56}$ reflects a smoother, more continuous interaction pattern, while $f_{78}$ incorporates sharp jumps and jagged edges, challenging the model's ability to capture more abrupt changes in relationships. This mix of interactions allows a comprehensive evaluation of the model's flexibility and accuracy in modeling diverse data patterns.

\subsubsection{Variable Selection} \label{subsec:var_select_syn}

We begin our experiment by performing variable selection as outlined in \cref{subsec:sparsity}, ensuring alignment with the sparsity requirement. Both synthetic datasets are processed using the same methodology. All neural network developments in this paper are implemented in Python using the \texttt{tensorflow} library.

The first step in this process is to select the significant main effects by creating an ensemble of 10 different Actuarial NAMs. Each model in the ensemble is trained with a different random seed in Keras, which results in varied initial model parameters and potentially distinct outputs. To account for the anticipated monotonic decreasing relationship with the response, we apply a lattice regression model to $X_3$ while the other covariates are modeled with fully-flexible neural networks. To manage computational costs, each subnetwork is designed with a shallow architecture comprising 2 hidden layers in a triangle-shaped structure, with 20 neurons in the first layer and 10 in the second. The hidden layers use the Leaky ReLU activation function. For the lattice regression model, we use a 1-D lattice with 10 vertices, equivalent to linear splines with 10 knots, enforcing a monotonic decreasing constraint. Batch Normalization layers in Keras are applied to standardize the mean of each shape function to 0. The final output layer uses an exponential activation function.

Each dataset, with 50,000 observations, is split into training, validation, and test sets with a 60:20:20 ratio. Each model in the ensemble is trained using the Adam optimizer, a maximum of 5000 epochs, and a batch size of 1000. We use a gamma negative log-likelihood as the objective function for optimization and apply early stopping with a patience of 10 to prevent overfitting. During this stage, we omit smoothness or marginal clarity penalties in the objective function, focusing solely on variable selection.

\begin{figure}[h]
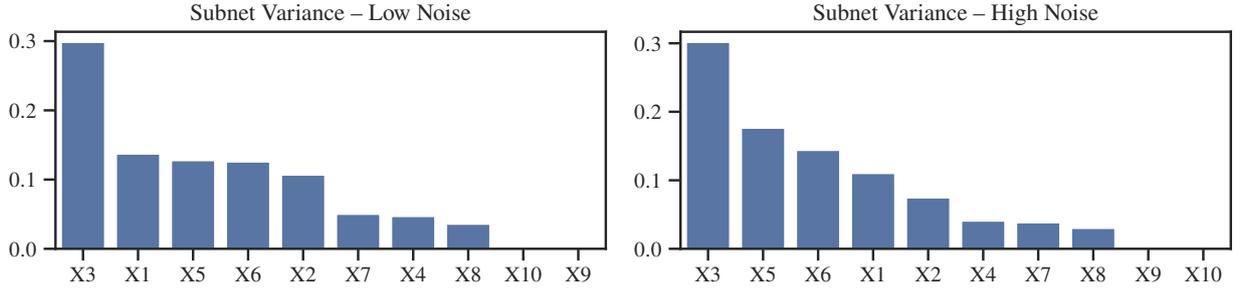

    \centering
    \begin{minipage}{0.5\linewidth}
        \centering
        \resizebox{\linewidth}{!}{\input{images/syn_low_main_selection.pgf}}
    \end{minipage}%
    \hfill
    \begin{minipage}{0.5\linewidth}
        \centering
        \resizebox{\linewidth}{!}{\input{images/syn_high_main_selection.pgf}}
    \end{minipage}
    \vspace{-1em}
    \caption{The average variances of the shape functions in an ensemble of 10 Actuarial NAMs.}
    \label{img:syn_main_selection}
\end{figure}

\Cref{img:syn_main_selection} shows the average variance of each shape function across the 10 models in the ensemble. Under both low-noise and high-noise conditions, the shape functions for $X_9$ and $X_{10}$ exhibit near-zero variance, significantly lower than that of the other covariates. In contrast, the important terms $X_1$ through $X_8$, used in the synthetic data generation, display distinctly higher variances. This demonstrates the ability of the Actuarial NAM to filter out noise and accurately identify the relevant features present in the ground truth. Consequently, the eight significant terms $X_1$ through $X_8$ are retained in the model, and a refined model containing only these terms is fitted with the same configurations discussed above. The validation loss from this model serves as the baseline for the next stage of our analysis.

Since we adopt a strong heredity constraint, a pairwise interaction term is only considered if both of its associated main effects were selected as important variables in the first step. With eight significant variables identified, there are 28 possible pairwise terms to evaluate. When each pairwise term is added to the model (which initially contains only the significant main effects), we use the same configurations as in the first step. However, the subnetwork or lattice model used to represent a pairwise term is more complex than those for the main effects to capture the additional intricacies of interactions. Specifically, we employ a deeper architecture with 10 hidden layers, starting with 100 neurons and linearly decreasing in size toward the output layer.

For pairwise effects requiring lattice models, inputs are first standardized as per \cref{eq:min_max}, using a piecewise-linear calibration layer from the \texttt{tensorflow-lattice} package. This layer scales the input to the appropriate range and enforces monotonicity when necessary. The preprocessed input is then fed into a lattice model with 8 vertices along each dimension, resulting in a total of 64 vertices to capture complex interaction patterns.

\begin{figure}[h]
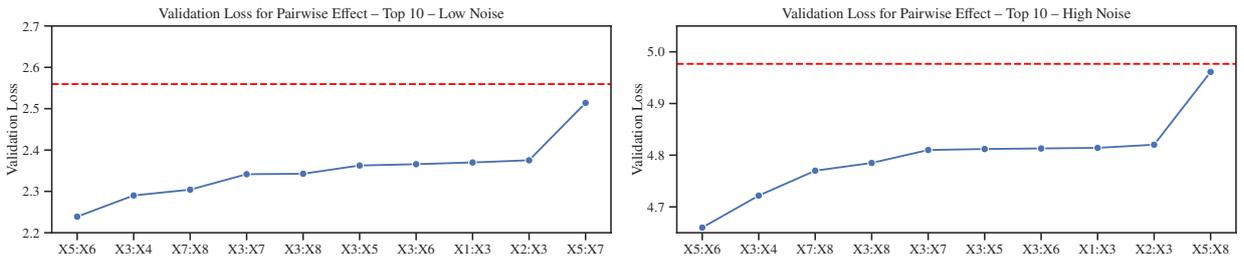

    \centering
    \begin{minipage}{0.5\linewidth}
        \centering
        \resizebox{\linewidth}{!}{\input{images/syn_low_int_selection.pgf}}
    \end{minipage}%
    \hfill
    \begin{minipage}{0.5\linewidth}
        \centering
        \resizebox{\linewidth}{!}{\input{images/syn_high_int_selection.pgf}}
    \end{minipage}
    \caption{Plots showing the top 10 pairwise effects that reduce the baseline validation loss the most. The red line shows the validation loss of the model with only significant main effects.}
    \label{img:syn_int_selection}
\end{figure}

\Cref{img:syn_int_selection} shows the 10 models that contain a single pairwise term alongside all significant main effects and reduce the baseline validation loss the most. Under both low-noise and high-noise scenarios, the top three pairwise terms are $X_3$:$X_4$, $X_5$:$X_6$, and $X_7$:$X_8$, which correspond to the true interactions in the ground truth. Our approach is to select terms that significantly improve validation loss without introducing excessive complexity, which can affect the computational cost. In the low-noise case, the top three terms—$X_3$:$X_4$, $X_5$:$X_6$, and $X_7$:$X_8$—are distinguishable and thus selected. For the high-noise case, we choose $X_3$:$X_4$, $X_5$:$X_6$, and $X_7$:$X_8$ but omit the $X_3$:$X_8$ term. This decision stems from the fact that the main effects of $X_3$ and $X_8$ are already represented in the selected pairwise terms, allowing them to be modeled indirectly. By limiting additional terms, we also mitigate unnecessary increases in computational cost. Overall, our experiments validate that the proposed variable selection process can highlight the critical terms (those consistent with the ground truth) while supporting sparsity and computational efficiency.

\subsubsection{Actuarial NAM --- Modeling Setup}

With the selected eight main effects and three pairwise effects, the next step is to perform hyperparameter tuning and fit a complete Actuarial NAM. We do not perform any preprocessing for the input covariates since they are all on the same scale, being generated from the same distribution $\mathcal{U}(-1, 1)$ as described in \cref{subsec:syn_data_mecha}. However, we need to remove some observations with extreme values in the response variable $Y$ to avoid skewing our analysis. We remove observations where $Y$ is beyond 1.5 times the interquartile range above the 75th percentile or below the 25th percentile \citep{tukey1977}. This results in 0.8\% of the data being removed in the low-noise scenario and 4\% in the high-noise scenario, out of a total of 50,000 observations in each dataset. We then split the remaining data into training, validation, and test sets with a 60:20:20 ratio.

Since the output is expected to have a monotonically decreasing relationship with $X_3$, we use lattice regression models to model the main effect of $X_3$ and the interaction effect between $X_3$ and $X_4$. All other shape functions are modeled using feedforward neural networks. Unlike the configurations used in the variable selection process, here we can afford to fit a deep subnetwork for each shape function, as we are building our final model. We adopt Leaky ReLU as the activation function in the hidden layers of the subnetworks and an exponential activation function for the final output layer.

In this stage, the model is trained with smoothness and marginal clarity penalties in the objective function, together with the Gamma negative log-likelihood. Smoothness requirements are applied to $X_1$ and $X_2$ to ensure that our model can capture the underlying smooth curves in the ground truth. We generate 1,000 uniformly spaced points from the domain of each of $X_1$ and $X_2$ to facilitate the implementation of the smoothness requirement as given in \cref{subsec:smoothness}. The prediction of our model is clipped between $10^{-7}$ and $10^{30}$ to prevent numerical overflow issues during training.

For hyperparameter optimization, eight different hyperparameters are tuned, with the details given in \ref{append:syn_ANAM_tuning}. We try 20 different combinations of the hyperparameters by fitting each model on the training set and choosing the one that has the lowest Gamma negative log-likelihood on the validation set. The model is trained with the Adam optimizer, a maximum of 5,000 epochs, and a batch size of 5,000. Early stopping with a patience of 60 epochs is applied to avoid overfitting. All neural network parameters in the Actuarial NAM are initialized using the default Glorot initialization \citep{glorot2010} of TensorFlow. For lattice regression models, the parameters are initialized using the default method of TensorFlow Lattice to form a linear function with positive and equal coefficients for monotonic dimensions and zero coefficients for other dimensions. We use 10 iterations for Dykstra's projection algorithm, which is the default in TensorFlow Lattice for projecting the parameters onto the feasible sets.

\subsubsection{Model Interpretability}

\begin{figure}[h]
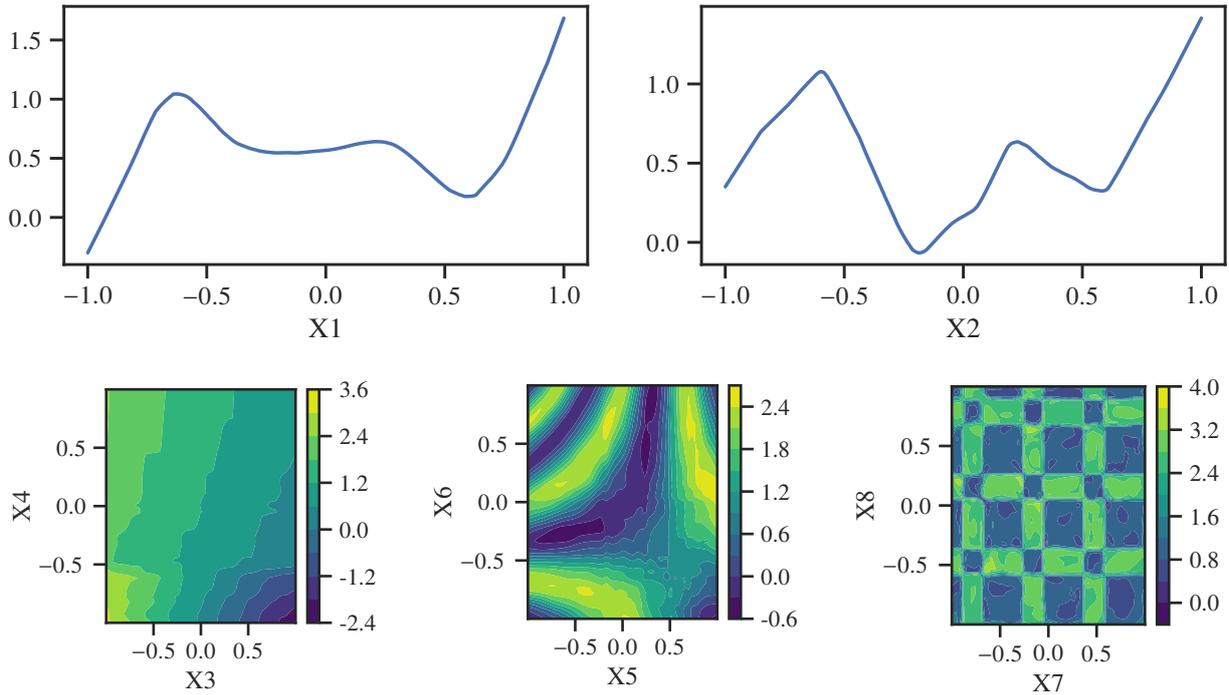

    \centering
    % First row with 2 images
    \begin{minipage}{0.49\linewidth}
        \centering
        \resizebox{\linewidth}{!}{\input{images/syn_low_X1.pgf}}
    \end{minipage}
    \hfill
    \begin{minipage}{0.49\linewidth}
        \centering
        \resizebox{\linewidth}{!}{\input{images/syn_low_X2.pgf}}
    \end{minipage}
    % Second row with 3 images
    \begin{minipage}{0.32\linewidth}
        \centering
        \resizebox{\linewidth}{!}{\input{images/syn_low_X3_X4.pgf}}
    \end{minipage}
    \hfill
    \begin{minipage}{0.32\linewidth}
        \centering
        \resizebox{\linewidth}{!}{\input{images/syn_low_X5_X6.pgf}}
    \end{minipage}
    \hfill
    \begin{minipage}{0.32\linewidth}
        \centering
        \resizebox{\linewidth}{!}{\input{images/syn_low_X7_X8.pgf}}
    \end{minipage}
    \caption{Covariate effects of Actuarial NAM --- Low noise}
    \label{fig:syn_low_anam}
\end{figure}

\begin{figure}[h]
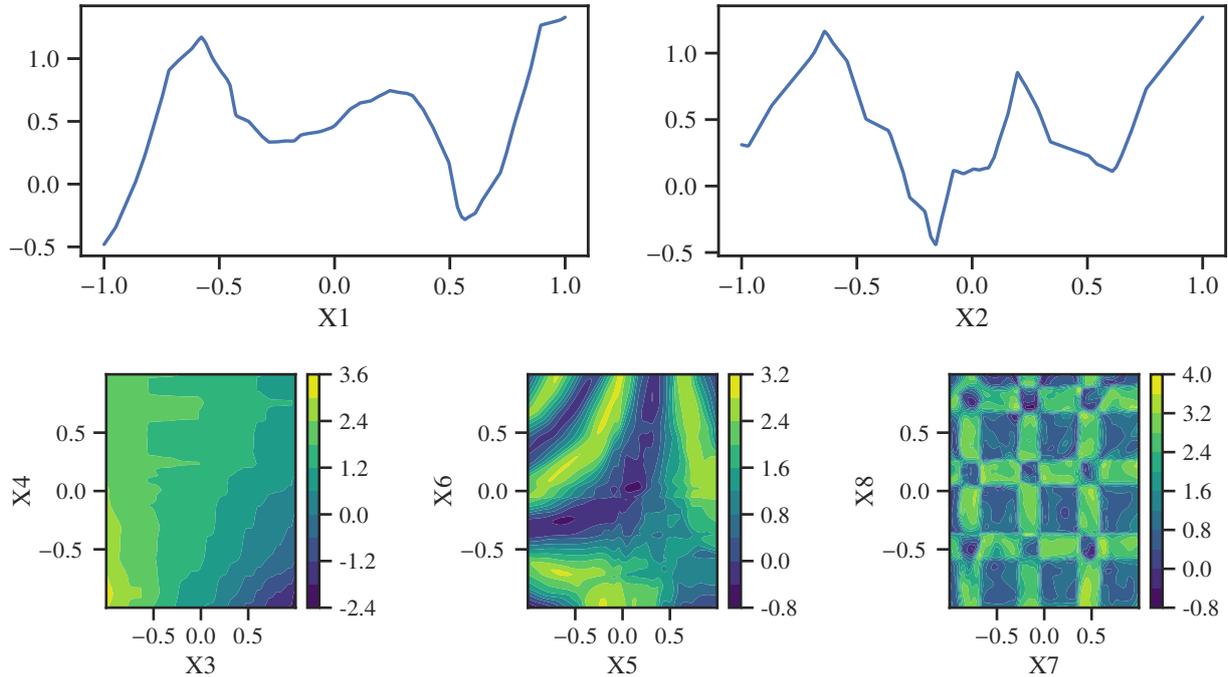

    \centering
    % First row with 2 images
    \begin{minipage}{0.49\linewidth}
        \centering
        \resizebox{\linewidth}{!}{\input{images/syn_high_X1.pgf}}
    \end{minipage}
    \hfill
    \begin{minipage}{0.49\linewidth}
        \centering
        \resizebox{\linewidth}{!}{\input{images/syn_high_X2.pgf}}
    \end{minipage}
    % Second row with 3 images
    \begin{minipage}{0.32\linewidth}
        \centering
        \resizebox{\linewidth}{!}{\input{images/syn_high_X3_X4.pgf}}
    \end{minipage}
    \hfill
    \begin{minipage}{0.32\linewidth}
        \centering
        \resizebox{\linewidth}{!}{\input{images/syn_high_X5_X6.pgf}}
    \end{minipage}
    \hfill
    \begin{minipage}{0.32\linewidth}
        \centering
        \resizebox{\linewidth}{!}{\input{images/syn_high_X7_X8.pgf}}
    \end{minipage}
    \caption{Covariate effects of Actuarial NAM --- High noise}
    \label{fig:syn_high_anam}
\end{figure}

We first evaluate the interpretability of the results and Actuarial NAM's ability to recover the ground truth under different levels of noise. As previously discussed, the model is interpretable because we can obtain an exact description of the internal logic by visualizing the shape functions produced by each subnetwork or lattice model.  \Cref{fig:syn_low_anam,fig:syn_high_anam} provide the covariate effects of the best models obtained under low-noise and high-noise scenarios, respectively. Looking at these figures and the ground truth in \Cref{img:ground_truth_main,img:ground_truth_interaction}, we can see that Actuarial NAM can capture the patterns that exist in the ground truth, with the overall relationships represented fairly accurately.

In the low-noise case, the main effects of $X_1$ and $X_2$ are well represented, though Actuarial NAM may oversmooth these relationships, missing certain turning points—especially for $X_2$'s main effect. This trade-off between interpretability and accuracy can be controlled by adjusting the smoothness penalty’s regularization strength. The monotonic decreasing relationship is enforced, as seen in the $X_3$ and $X_4$ interaction heatmap. The pairwise effects are also captured very well, although they are more jagged than the underlying ground truth. Abrupt changes in $X_7:X_8$ interactions are accurately represented, and the smooth characteristics of $X_5:X_6$ interactions are adequately captured.

In the high-noise case, the relationships are less accurately captured, though the general patterns are sufficiently represented and clearly seen in the plots. Due to the high level of noise in the data, Actuarial NAM overfitted slightly, and the smoothness constraint wasn't strong enough. As a result, the line charts for the main effects of $X_1$ and $X_2$ are slightly more jagged than we expect. Despite this, the high-level patterns are still correct. The same goes for the pairwise interaction effects. Monotonicity is well captured for $X_3$, though the heatmaps for all three pairwise terms are noisier than the underlying ground truth. This is not unexpected, given the higher noise level in these synthetic datasets. Overall, Actuarial NAM has demonstrated the ability to recover and capture the ground truth sufficiently well under both simulated scenarios.

\subsubsection{Prediction Accuracy}
To evaluate prediction accuracy, we compare Actuarial NAM with the models listed in \cref{subsec:model_eval}. Hyperparameter tuning is conducted for all candidate models using the validation set. After tuning, we assess the predictive performance of each model on the test set.

For GLM, GBM, LocalGLMnet, and Neural Networks, we use all 10 predictors, from $X_1$ to $X_{10}$, to generate predictions. EBM, a tree-based version of GAM that models both main and interaction effects, is implemented using the open-source Python package \texttt{interpret}, which includes an internal variable selection process to identify important pairwise terms, while still using all variables for main effects. For GAM, we use the \texttt{pygam} package with the default cubic splines for modeling main effects and multi-dimensional splines with tensor product bases for pairwise interactions, as outlined in \citet{hastie2017}. GAM undergoes a variable selection process consistent with Actuarial NAM to ensure comparability. This approach places all GAM-based models on equal footing, allowing a fair assessment of each method's effectiveness. Notably, GAM with splines selects the same 8 main effects and 3 interactions as Actuarial NAM in both low-noise and high-noise scenarios. All models are optimized to minimize the Gamma negative log-likelihood objective function.

\Cref{tbl:syn_data} presents the predictive performance of all models on the test set. Actuarial NAM demonstrates superior performance in both scenarios, achieving the lowest NLL and RMSE values while offering competitive performance with black-box models in terms of MAE. These results underscore Actuarial NAM's strong predictive power, offering enhanced prediction accuracy compared to traditional insurance pricing models like GLM and GAM, while also achieving competitive performance with black-box machine learning models and providing fully transparent results and internal logic.

One notable observation from \cref{tbl:syn_data} is that GAM and EBM yield high prediction accuracy, even when compared to black-box models like Neural Networks or GBM. This is likely due to the additive nature of the ground truth, which comprises only main and pairwise interaction effects—precisely the structure GAM and EBM are designed to capture. Additionally, LocalGLMnet and Neural Networks perform relatively poorly. This may be attributed to the lack of an effective variable selection process, which could lead to overfitting. Without proper selection, these models may struggle to generalize effectively, as they are likely to fit noise along with the signal. This highlights the importance of robust variable selection in complex models, particularly when capturing interpretable patterns is a primary goal. Overall, Actuarial NAM’s balance of interpretability and predictive performance positions it as a valuable approach for accurate and insightful insurance pricing.

\begin{table}[h!]
    \centering
    \begin{tabular}{@{}lccc@{\hspace{2em}}ccc@{}}
        \toprule
        & \multicolumn{3}{c}{Low Noise} & \multicolumn{3}{c}{High Noise} \\
        \cmidrule(lr){2-4} \cmidrule(lr){5-7}
        Model & NLL & RMSE & MAE & NLL & RMSE & MAE \\
        \midrule
        GLM           & 2.64 & 2450.76 & 1097.36 & 5.17 & 4832.31 & 1380.31 \\
        GAM           & 1.69 & 2029.96 & 721.05  & 4.28 & 4648.41 & 1129.99 \\
        EBM           & 1.80 & 2189.64 & 780.22  & 4.45 & 4917.47 & 1255.79 \\
        LocalGLMnet   & 2.10  & 2327.84 & 966.57  & 4.72 & 4907.89 & 1466.25 \\
        GBM           & 1.72 & 2033.16 & \textbf{664.89}  & 4.57 & 4686.33 & \textbf{1044.83} \\
        Neural Nets   & 2.16 & 2303.25 & 919.96  & 4.86 & 4794.13 & 1466.47 \\
        Actuarial NAM & \textbf{1.65} & \textbf{2020.66} & 698.90  & \textbf{4.25} & \textbf{4632.97} & 1167.38 \\
        \bottomrule
    \end{tabular}
    \caption{Comparison of predictive performance on the test set under different noise levels. The best values are bolded.}
    \label{tbl:syn_data}
\end{table}

\subsection{Real Dataset Illustration} \label{subsec:real_data}
\subsubsection{Data Information and Modeling Setup}
In this section, we use a Belgian motor third-party liability dataset from the \texttt{CASdatasets} collection. This dataset is collected from an undisclosed insurer and contains information on 163,212 unique policies. Each row provides details on the number of claims and the total claim amount for each policy during a period of exposure to risk, expressed as the fraction of the year during which the policyholder was exposed to the risk of filing a claim. Additionally, the dataset includes 12 columns capturing policy information—6 categorical and 6 continuous variables—that can be used as predictors for our model. A summary of all columns in the dataset is provided in \ref{append:var_list}.

Our focus is on predicting the claim frequency, represented by the variable \texttt{nclaims}. We are particularly interested in this target variable because we need to have a monotonic relationship to test our model's ability to enforce the monotonicity requirement, and the dataset includes a \texttt{bm} (Bonus--Malus) column, which is expected to have a monotonically increasing relationship with claim frequency.

We preprocess the data by removing irrelevant columns, including \texttt{id}, \texttt{claim}, \texttt{amount}, \texttt{average}, and \texttt{postcode}. The variable \texttt{postcode} is removed because, according to the data description, it contains information that is already represented by the two continuous variables \texttt{lat} (latitude) and \texttt{long} (longitude). We standardize all six continuous variables to have zero mean and a standard deviation of one while encoding the categorical variables using one-hot encoding. The target variable \texttt{nclaims} contains values ranging from 0 to 5, so there are no extreme outliers to manage. Additionally, we incorporate the exposure information (the \texttt{expo} column) in our modeling process to account for the fact that different policies have different exposures to risk.

Since the target variable \texttt{nclaims} represents the number of claims for an insurance policy, a natural choice is to assume a Poisson distribution for modeling. This approach aligns with the methodology adopted in \citet{denuit2004} on the same dataset. We use a log link function to relate the expected claim frequency to the predictors in our Actuarial NAM. Additionally, the exposure information is applied as an offset term in our modeling framework to adjust for varying exposure times among policies. The mathematical formulation of our model can thus be expressed as:
\begin{equation}
    \log(\mu) = \beta + \boldsymbol{\alpha}_{1}\sum_{i = 1}^ps_i(X_i) + \boldsymbol{\alpha}_{2}\sum_{j = 1}^p\sum_{\substack{k=1 \\ k \ne j}}^j h_{jk}(X_j, X_k) + \log(\lambda)
\end{equation}
where $\beta$, $\boldsymbol{\alpha}_{1}$, and $\boldsymbol{\alpha}_{2}$ respectively denote the bias term and the vector of weights connecting the output of the univariate and bivariate shape functions with the output neuron of the models; $p$ is the number of covariates; $s_i(X_i)$ and $h_{jk}(X_j,X_k)$ respectively represent the shape functions of univariate and bivariate effects; and $\lambda$ is the exposure to risk.

The above formulation presents the general case, incorporating all covariates and potential pairwise interactions. However, in this analysis, we will perform a variable selection process to comply with the sparsity requirement, ensuring that the model remains interpretable and avoids overfitting. The purpose of presenting the above formulation is to demonstrate how the offset term is incorporated into the modeling framework.

\subsubsection{Variable Selection}
The data is divided into training, validation, and test sets with a 60:20:20 split, consistent with our approach for synthetic datasets. We apply the same model configurations as in the synthetic dataset analysis for the initial variable selection phase. The \texttt{bm} variable is modeled with lattice regression, using 10 vertices to enforce a monotonically increasing relationship with the output. All other main effects are modeled with a shallow neural network comprising two hidden layers with 20 and 10 neurons, respectively.

We create an ensemble of 10 models, each optimized with the RMSprop optimizer to minimize Poisson deviance. Training is performed over a maximum of 5,000 epochs with a batch size of 5,000, and early stopping with a patience of 10 epochs is applied to mitigate overfitting. This approach helps ensure that the model accurately captures the underlying relationships in the data while remaining resilient against noise.

\begin{figure}[h]
    \centering
    \resizebox{\linewidth}{!}{\input{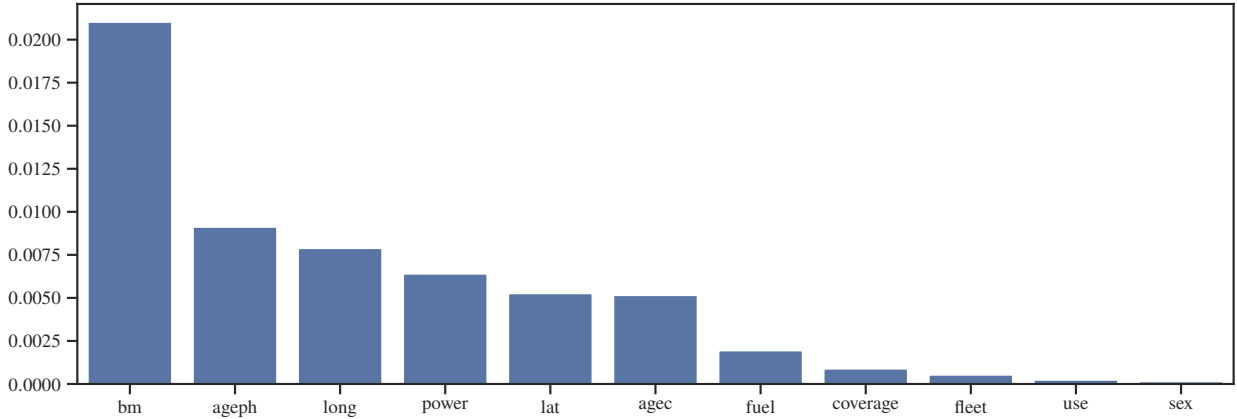}}
    \vspace{-2em}
    \caption{The average variances of the main effects' shape functions in an ensemble of 10 Actuarial NAMs.}
    \label{fig:real_main_selection}
\end{figure}

The average variance of each shape function across an ensemble of 10 different Actuarial NAM models is illustrated in \cref{fig:real_main_selection}. From this figure, we observe that the variables \texttt{coverage}, \texttt{fleet}, \texttt{use}, and \texttt{sex} exhibit near-zero variances, indicating limited relevance in explaining the response variation. As these variables contribute minimally to the model’s predictive power, they are excluded from further analysis. Additionally, although the \texttt{fuel} variable shows a relatively low variance among the remaining terms, it still exhibits a non-negligible contribution and is therefore included in the model. This leaves us with a total of 7 significant main effects, which we use to fit an Actuarial NAM and establish a baseline validation loss for selecting the pairwise terms.

For selecting significant pairwise interaction effects, we consider 21 possible pairs. Each pairwise interaction is modeled with a neural network employing a triangular hidden layer structure, starting with 100 neurons in the first layer and linearly decreasing over 10 layers. For pairs involving the \texttt{bm} variable, a lattice model with 8 vertices along each dimension is applied to enforce the expected monotonic relationship. The input to this lattice model is scaled to the appropriate range using a piecewise linear calibrator with 20 knots, ensuring consistency in input scaling and enabling accurate function approximation.

\begin{figure}[h]
    \centering
    \resizebox{\linewidth}{!}{\input{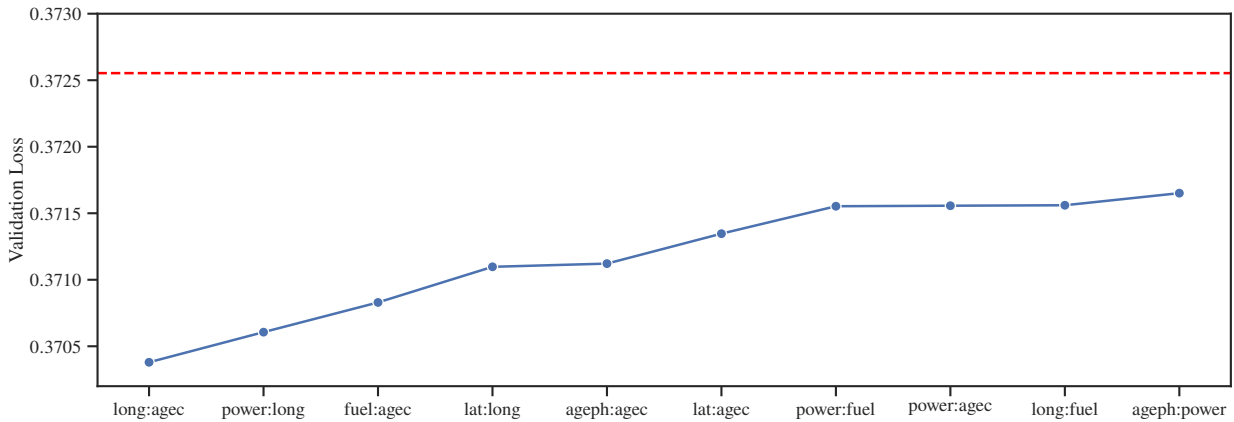}}
    \vspace{-2em}
    \caption{Plots showing the top 10 pairwise effects that reduce the baseline validation loss the most. The red line shows the validation loss of the model with only significant main effects.}
    \label{fig:real_int_selection}
\end{figure}

The results of the pairwise effect selection are displayed in \cref{fig:real_int_selection}, which highlights the 10 pairwise effects that most significantly reduce the baseline validation loss when added to a model containing only the significant main effects. Based on this analysis, we select the top 5 pairwise interactions: \texttt{long}:\texttt{agec}, \texttt{power}:\texttt{long}, \texttt{fuel}:\texttt{agec}, \texttt{lat}:\texttt{long}, and \texttt{ageph}:\texttt{agec}. These pairs are chosen not only because they yield the greatest reduction in validation loss, but also because they encompass all main effects present in the top 10 pairwise effects. Including these 5 pairs allows the model to indirectly capture interactions represented by other combinations in the top 10, thus reducing the risk of overfitting and keeping computational costs manageable by limiting the number of terms.

\subsubsection{Model Interpretability}

With the selected 7 main effects and 5 pairwise interaction effects, we proceed to fit the final Actuarial NAM model, incorporating both marginal clarity and smoothness penalties in the objective function. A smoothness constraint is applied to all 6 continuous variables in the selected main effects. To implement this, we generate 1,000 uniformly spaced points from each variable’s domain and use the finite differences method, as discussed in \cref{subsec:smoothness}, to penalize excessive jaggedness in each shape function. As with the synthetic datasets, we tune 8 hyperparameters in this model, detailed in \ref{append:syn_ANAM_tuning}. We test 150 combinations of these hyperparameters and select the configuration that yields the lowest Poisson deviance on the validation set. Each model is trained for up to 5,000 epochs with a batch size of 10,000. Leaky ReLU is used as the activation function for the hidden layers of each subnetwork, while an exponential activation function is applied in the output layer.

\begin{figure}[h]
    \centering
    \resizebox{\linewidth}{!}{\input{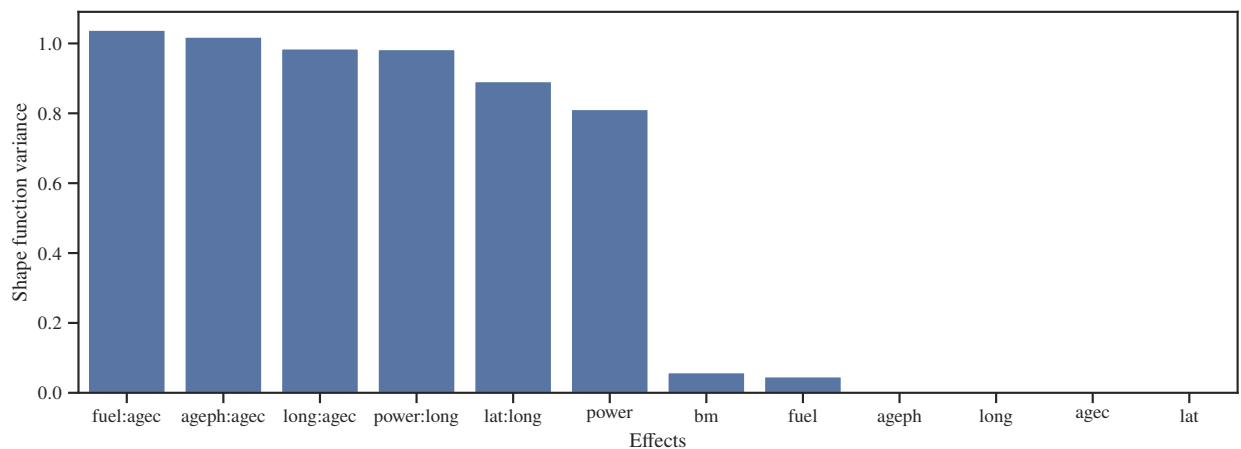}}
    \vspace{-2em}
    \caption{Quantification of the importance of each effect used by Actuarial NAM.}
    \label{fig:real_var_importance}
\end{figure}

To quantify the significance of each main and pairwise term in Actuarial NAM, we assess the variance of each shape function, as shown in \cref{fig:real_var_importance}. This analysis reveals that with the inclusion of pairwise terms, four main terms—\texttt{ageph}, \texttt{long}, \texttt{agec}, and \texttt{lat}—are less significant individually in explaining the response variation. This reduced significance likely results from the enhanced explanatory power of these variables when evaluated together in pairwise terms. For instance, latitude and longitude are more informative as a pair, providing more meaningful geographic information than they do individually, as they together represent a precise geographical coordinate.

\Cref{fig:real_anam} displays all 12 shape functions produced by the model. Each main effect can be interpreted as the average influence of a variable on claim frequency, while each pairwise interaction refines this effect based on the level of another variable. The model successfully enforces the monotonicity requirement, as seen in the shape function for \texttt{bm}, which maintains a consistent, monotonically increasing trend. The smoothness requirement is also well-implemented, with main effect shape functions appearing smooth and free of unnecessary jaggedness. 

By visualizing the shape functions, we gain clear insight into how the model arrives at specific predictions and the relationships it has learned from the data. For instance, the model captures the trend that higher vehicle or policyholder age is associated with a lower expected claim frequency. This level of interpretability—allowing us to examine the model’s internal logic and assess requirements like smoothness and monotonicity—sets our model apart from other neural network architectures, which often lack this level of transparency and control.

\begin{figure}[!htbp]
    \centering
    % First row with 3 images
    \begin{minipage}{0.32\linewidth}
        \centering
        \resizebox{\linewidth}{!}{\input{images/real_shape_agec.pgf}}
    \end{minipage}
    \hfill
    \begin{minipage}{0.32\linewidth}
        \centering
        \resizebox{\linewidth}{!}{\input{images/real_shape_ageph.pgf}}
    \end{minipage}
    \hfill
    \begin{minipage}{0.32\linewidth}
        \centering
        \resizebox{\linewidth}{!}{\input{images/real_shape_bm.pgf}}
    \end{minipage}
    
    % Second row with 3 images
    \begin{minipage}{0.32\linewidth}
        \centering
        \resizebox{\linewidth}{!}{\input{images/real_shape_fuel.pgf}}
    \end{minipage}
    \hfill
    \begin{minipage}{0.32\linewidth}
        \centering
        \resizebox{\linewidth}{!}{\input{images/real_shape_lat.pgf}}
    \end{minipage}
    \hfill
    \begin{minipage}{0.32\linewidth}
        \centering
        \resizebox{\linewidth}{!}{\input{images/real_shape_long.pgf}}
    \end{minipage}
    
    % Third row with 3 images
    \begin{minipage}{0.32\linewidth}
        \centering
        \resizebox{\linewidth}{!}{\input{images/real_shape_power.pgf}}
    \end{minipage}
    \hfill
    \begin{minipage}{0.32\linewidth}
        \centering
        \resizebox{\linewidth}{!}{\input{images/real_shape_fuel_agec.pgf}}
    \end{minipage}
    \hfill
    \begin{minipage}{0.32\linewidth}
        \centering
        \resizebox{\linewidth}{!}{\input{images/real_shape_ageph_agec.pgf}}
    \end{minipage}
    
    % Fourth row with 3 images
    \begin{minipage}{0.32\linewidth}
        \centering
        \resizebox{\linewidth}{!}{\input{images/real_shape_lat_long.pgf}}
    \end{minipage}
    \hfill
    \begin{minipage}{0.32\linewidth}
        \centering
        \resizebox{\linewidth}{!}{\input{images/real_shape_long_agec.pgf}}
    \end{minipage}
    \hfill
    \begin{minipage}{0.32\linewidth}
        \centering
        \resizebox{\linewidth}{!}{\input{images/real_shape_power_long.pgf}}
    \end{minipage}    
    \caption{Shape functions generated by Actuarial NAM on the real dataset.}
    \label{fig:real_anam}
\end{figure}

\subsubsection{Prediction Accuracy}
We conduct hyperparameter tuning across all candidate models to optimize predictive performance and ensure a fair comparison. Feature selection is applied to the spline-based GAM, as the selection method used for Actuarial NAM is also compatible with GAM. All other models utilize the full set of 11 input variables, with EBM conducting a selection of significant pairwise effects.

\Cref{tbl:real_data} shows the predictive performances of Actuarial NAM and all other candidate models on the test set. We observe that Actuarial NAM outperforms all other models in all chosen metrics. LocalGLMnet, a popular neural network architecture designed for interpretable actuarial modeling, also performs well on this dataset, offering competitive performance to the black-box models. The fact that Actuarial NAM can offer better prediction accuracy while providing transparent internal logic underscores its advantage in bridging the gap between interpretability and high predictive power.

\begin{table}[h!]
    \centering
    \begin{tabular}{@{}lccc@{}}
        \toprule
        & \multicolumn{3}{c}{\small (in units of $10^{-2}$)} \\
        Model & NLL & RMSE & MAE \\
        \midrule
        GLM           & 38.03 & 36.86 & 21.77 \\
        GAM           & 38.40 & 37.29 & 23.88 \\
        EBM           & 38.01 & 36.91 & 22.50 \\
        LocalGLMnet   & 37.97 & 36.83 & 22.17 \\
        GBM           & 37.98 & 37.06 & 22.44 \\
        Neural Nets   & 37.95 & 36.81 & 21.72 \\
        Actuarial NAM & \textbf{37.88} & \textbf{36.79} & \textbf{21.57} \\
        \bottomrule
    \end{tabular}
    \caption{Comparison of predictive performance on the test set for the real dataset. The best values are bolded.}
    \label{tbl:real_data}
\end{table}

\section{Conclusion} \label{sec:con}

Deep learning models, while powerful, lack transparency, posing challenges in high-stakes decision-making like insurance pricing. In this paper, we tackle the gap in the actuarial literature concerning the development of an inherently interpretable pricing model that incorporates neural network components by making two main contributions:
\begin{enumerate}
    \item \textbf{Establish an interpretability framework in insurance pricing with a concrete definition and mathematical formulation of interpretable pricing models clearly defined}. We formulate the definition by recognizing that interpretability is a domain-specific concept, and a model can only be considered interpretable when exact descriptions of the cause and effect within the model are available and understandable to humans. We also identify six interpretability and practical requirements relevant to insurance pricing, including the transparency of main and interaction effects, the ability to quantify variables' importance, sparsity, smoothness, and monotonicity. A mathematical framework for interpretability is created by extending the approach in \citet{rudin2022}, where the interpretability and practical requirements can be implemented via either penalty terms in the objective function or architectural restrictions.
    \item \textbf{Develop Actuarial NAM, an inherently interpretable pricing model with neural network components that satisfies all key interpretability requirements specific to general insurance pricing}. Our modeling approach enhances the Neural Additive Model, originally proposed by \citet{agarwal2021}, by adding pairwise interaction effects and making various adjustments to the network architecture and training process to incorporate the prespecified interpretability requirements particularly suitable for actuarial applications in pricing. Specifically, we adopt lattice regression to enforce monotonicity, introduce penalty terms in the optimization algorithm to promote smoothness and design a three-stage process for effective variable selection. Under this modeling framework, interpretability is attained through visualization of line charts or heatmaps summarizing the shape functions, alongside the guarantee that the model output aligns with domain requirements.
\end{enumerate}

Experiments on both synthetic and real datasets have shown that Actuarial NAM can outperform classical insurance pricing and advanced machine learning models in most scenarios considered while providing results that are transparent and easy to interpret. The three-stage variable selection process can correctly identify the important terms under both low and high-noise simulated environments, while lattice regression and the inclusion of a smoothness penalty term ensures that the model output aligns with domain requirements. These results highlight the potential for commercial application of our model.

Limitations and further research: Despite its strong predictive performance, ANAM has several limitations that open avenues for future research. The model currently predicts only the conditional mean under a simple GLM-like variance assumption. Extending the model to a multi-output framework, modeling multiple distributional parameters or properties in a flexible way, would enable the adoption of more flexible distributional assumptions that fit well to the data. A potential way to enhance ANAM and simultaneously model multiple distributional properties is provided in \ref{append_B}. Additionally, while Actuarial NAM enforces full interpretability, exploring models that allow limited flexibility—such as boosting Actuarial NAM with a fully flexible neural network—may enhance predictive accuracy. A multi-task learning framework could also be explored to model claim frequency and severity simultaneously, leveraging shared learning signals for improved generalization. We leave the above for future work.

\section*{Data and Code}
The code and data used in this paper are available on GitHub at \url{https://github.com/tupho289/Actuarial-NAM}.

% \bibliographystyle{elsarticle-harv}
% \bibliography{ref}

\appendix
\section{Additional Details for Modeling} \label{append_A}
\subsection{Hyperparameter Tuning --- Actuarial NAM} \label{append:syn_ANAM_tuning}
Below are the eight hyperparameters of Actuarial NAM that are tuned in this paper's analyses.
\begin{itemize}
    \item The number of hidden layers for the main effects' subnetworks.
    \item The number of neurons in the first hidden layer of the main effects' subnetworks. We use a triangle-shaped architecture for the hidden layers, so the number of neurons decreases linearly from the first layer to 1 at the final layer.
    \item The number of hidden layers for the interaction effects' subnetworks.
    \item The number of neurons in the first hidden layer of the interaction effects' subnetworks.
    \item The number of key points used by the 1-D piecewise linear calibrator to preprocess the inputs to the lattice model.
    \item The number of vertices per dimension for the lattice model.
    \item The regularization strength for the marginal clarity constraint $\omega_{\text{mc}}$.
    \item The regularization strength for the smoothness constraint $\omega_{\text{smooth}}$.
\end{itemize}

\subsection{Variable list of the beMTPL97 dataset} \label{append:var_list}
\begin{table}[H]
    \centering
    \begin{tabular}{@{}lp{10cm}@{}}
        \toprule
        \textbf{Variable} & \textbf{Description} \\
        \midrule
        id         & A numeric for the policy number. \\
        expo       & A numeric for the exposure to risk. \\
        claim      & A factor indicating the occurrence of claims. \\
        nclaims    & A numeric for the claim number. \\
        amount     & A numeric for the aggregate claim amount. \\
        average    & A numeric for the average claim amount. \\
        coverage   & A factor for the insurance coverage level: ``TPL'' only third party liability, ``TPL+'' TPL + limited material damage, ``TPL++'' TPL + comprehensive material damage. \\
        ageph      & A numeric for the policyholder age. \\
        sex        & A factor for the policyholder gender: ``female'', ``male''. \\
        bm         & An integer for the level occupied in the former compulsory Belgian bonus-malus scale. From 0 to 22, a higher level indicates a worse claim history.. \\
        power      & A numeric for the Horsepower of the vehicle in kilowatt. \\
        agec       & A numeric for the Age of the vehicle in years. \\
        fuel       & A factor for the Type of fuel of the vehicle: ``gasoline'' or ``diesel''. \\
        use        & A factor for the use of the vehicle: ``private'' or ``work''. \\
        fleet      & An integer indicating if the vehicle is part of a fleet: 1 or 0. \\
        postcode   & The postal code of the policyholder. \\
        long       & A numeric for the longitude coordinate of the center of the municipality where the policyholder resides. \\
        lat        & A numeric for the latitude coordinate of the center of the municipality where the policyholder resides. \\
        \bottomrule
    \end{tabular}
\end{table}

\section{ANAM with Flexible Distributional Regression} \label{append_B}
Let $p$ be the number of input factors and $\boldsymbol{\lambda} = [\lambda_1, \lambda_2, \dots, \lambda_q]^\top$ be the $q$-dimensional output vector of the model ($p, q \in \mathbb{N}$). Further, let $s_{oi}(X_i)$ denote the function learned by a subnetwork capturing the main effect of input variable $X_i$ on the $o$-th output, and let $h_{ojk}(X_j, X_k)$ be the function learned by a subnetwork representing the pairwise interaction effect between variables $X_j$ and $X_k$ ($1 \leq i,j,k \leq p$, $j \neq k$) for the $o$-th output. The Actuarial Neural Additive Model with multiple outputs can be expressed as:
\begin{equation}
    \label{eq:proposed_model}
    \left\{ \begin{array}{l}
  \lambda_1 = g_1(\beta_1 + \boldsymbol{\alpha}_{11}\sum_{i = 1}^ps_{1i}(X_i) + \boldsymbol{\alpha}_{12}\sum_{j = 1}^p\sum_{\substack{k=1 \\ k \ne j}}^j h_{1jk}(X_j, X_k)) \\
  \lambda_2 = g_2(\beta_2 + \boldsymbol{\alpha}_{21}\sum_{i = 1}^ps_{2i}(X_i) + \boldsymbol{\alpha}_{22}\sum_{j = 1}^p\sum_{\substack{k=1 \\ k \ne j}}^j h_{2jk}(X_j, X_k)) \\
  \\[-2ex] % Adjust spacing as needed
  \cdots \\
  \\[-2ex] % Adjust spacing as needed
  \lambda_q = g_q(\beta_q + \boldsymbol{\alpha}_{q1}\sum_{i = 1}^ps_{qi}(X_i) + \boldsymbol{\alpha}_{q2}\sum_{j = 1}^p\sum_{\substack{k=1 \\ k \ne j}}^j h_{qjk}(X_j, X_k)) \\
\end{array} \right.
\end{equation}
where $g_o(\cdot)$, $\beta_o$, $\boldsymbol{\alpha}_{o1}$, and $\boldsymbol{\alpha}_{o2}$ respectively denote the activation function, the bias term, and vector of weights connecting the output of the univariate and bivariate subnetworks with the $o$-th output neuron of the models; $s_{oi}(X_i)$ and $h_{ojk}(X_j,X_k)$ respectively represent the shape functions of univariate and bivariate effects associated with the $o$-th output learned by the subnetworks.

This structure enables flexible distributional modeling by allowing each neuron in the output layer to represent a distinct distributional parameter or property. As each parameter ($\lambda_1, \lambda_2, \dots, \lambda_q$) has its dedicated subnetworks, applying various interpretability requirements described in section \Cref{sec:mf} becomes straightforward. This model can be considered as the Neural Additive Model for Location, Scale, and Shape (NAMLSS), resembling the approach described in \citet{rigby2005}. For instance, if we assume a Gamma distribution, the output neurons can represent the distribution's shape and scale parameters.

\section{Dykstra's Projection Algorithm for Monotonicity Requirements} \label{append_C}
\begin{algorithm} 
\begin{algorithmic}[1]
    \STATE \textbf{Input}: Lattice parameters $\boldsymbol{\lambda}_j$, maximum iterations $K_{\text{max}}$, convergence threshold $\delta$, feasible sets $\{\mathcal{D}_1,\dotsc,\mathcal{D}_{v_j}\}$
    \STATE \textbf{Output}: Projected parameters $\boldsymbol{\lambda}_j^{*}$
    \STATE Initialize $\boldsymbol{\lambda}_j^{(0)} \leftarrow \boldsymbol{\lambda}_j$, auxiliary variables $\boldsymbol{p}_i^{(0)} \leftarrow \boldsymbol{0} \, \forall \,i \in \{1,2,\dotsc,v_j\}$, $k \leftarrow 0$
    \REPEAT
        \FOR{$i = 1, 2, \dots, v_j$}
            \STATE Compute $\boldsymbol{s}_i^{(k)} \leftarrow \boldsymbol{\lambda}^{(k)}_j + \boldsymbol{p}_i^{(k)}$
            \STATE Project $\boldsymbol{s}_i^{(k)}$ onto feasible set $\mathcal{D}_i$: $\boldsymbol{r}_i^{(k)} \leftarrow \operatorname{Proj}_{\mathcal{D}_i}(\boldsymbol{s}_i^{(k)})$
            \STATE Update auxiliary variable: $\boldsymbol{p}_i^{(k+1)} \leftarrow \boldsymbol{s}_i^{(k)} - \boldsymbol{r}_i^{(k)}$
            \STATE Update parameters: $\boldsymbol{\lambda}^{(k+1)}_j \leftarrow \boldsymbol{r}_i^{(k)}$
        \ENDFOR
        \STATE $k \leftarrow k + 1$
    \UNTIL{$||\boldsymbol{\lambda}_j^{(k)} - \boldsymbol{\lambda}_j^{(k-1)}|| < \delta$ or $k \geq K_{\text{max}}$}
    \STATE \textbf{return} $\boldsymbol{\lambda}_j^{*} \leftarrow \boldsymbol{\lambda}_j^{(k)}$
\end{algorithmic}
\end{algorithm}

\end{document}